\documentclass[10pt,twocolumn,letterpaper]{article}

\usepackage{cvpr}              %
\usepackage{makecell}
\usepackage{graphicx}
\usepackage{subcaption}
\usepackage{comment}
\usepackage{multirow}
\usepackage{array}
\usepackage{booktabs}
\usepackage{amsmath}
\usepackage{xcolor}
\usepackage{float}
\usepackage{colortbl}

\newcommand{\scorename}{OmniScore\xspace}

\newcommand{\method}{VideoDPO\xspace}

\newcommand{\GR}[0]{\cellcolor{gray!25}}
\definecolor{darkgreen}{rgb}{0.0, 0.5, 0.0}

\usepackage{lipsum}

\definecolor{cvprblue}{rgb}{0.21,0.49,0.74}
\usepackage[pagebackref,breaklinks,colorlinks,allcolors=cvprblue]{hyperref}

\makeatletter\renewcommand\paragraph{\@startsection{paragraph}{4}{\z@}
{.2em \@plus1ex \@minus.2ex}{-.5em}{\normalfont\normalsize\bfseries}}\makeatother

\def\paperID{5843} %
\def\confName{CVPR}
\def\confYear{2025}

\title{VideoDPO: Omni-Preference Alignment for Video Diffusion Generation}

\author{Runtao Liu$^1$\thanks{~Equal Contribution.} \quad Haoyu Wu$^2$\footnotemark[1] \quad Ziqiang Zheng$^1$ \quad Chen Wei$^3$ \\Yingqing He$^1$ \quad Renjie Pi$^1$ \quad Qifeng Chen$^1$\\
$^1$HKUST \quad $^2$Renmin University of China \quad $^3$Johns Hopkins University\\
{\tt\small rliuay@connect.ust.hk \quad haoyuwu556@ruc.edu.cn}
}

\begin{document}
\maketitle
\begin{abstract}

Recent progress in generative diffusion models has greatly advanced text-to-video generation. While text-to-video models trained on large-scale, diverse datasets can produce varied outputs, these generations often deviate from user preferences, highlighting the need for preference alignment on pre-trained models. Although Direct Preference Optimization (DPO)~\cite{rafailov2023DPO} has demonstrated significant improvements in language and image generation~\cite{wallace2023diffusionDPO}, we pioneer its adaptation to \emph{video} diffusion models and propose a \emph{VideoDPO} pipeline by making several key adjustments. Unlike previous image alignment methods that focus solely on either (\emph{i}) visual quality or (\emph{ii}) semantic alignment between text and videos, we comprehensively consider both dimensions and construct a preference score accordingly, which we term the \scorename. We design a pipeline to automatically collect preference pair data based on the proposed \scorename and discover that re-weighting these pairs based on the score significantly impacts overall preference alignment. Our experiments demonstrate substantial improvements in both visual quality and semantic alignment, ensuring that no preference aspect is neglected. 
Code and data are available at \url{https://videodpo.github.io/}. 

\end {abstract}
    
\section{Introduction}

\begin{figure*}[htbp]
    \centering

    \resizebox{1.0\textwidth}{!}{
    \begin{tabular}{@{}c@{}c@{}cc@{}c@{}c@{}}
        \includegraphics[width=75px, height=55px]{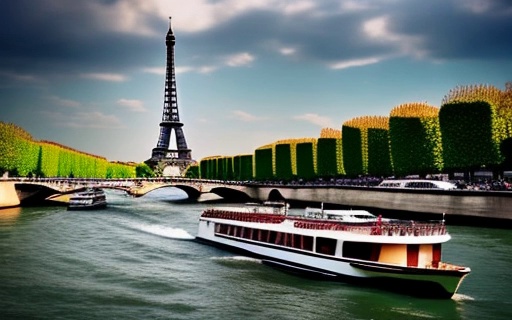} &
        \includegraphics[width=75px, height=55px]{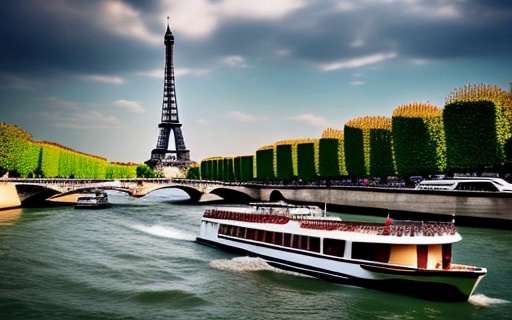} & 
        \includegraphics[width=75px, height=55px]{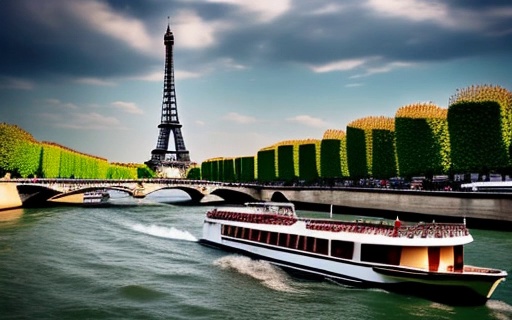} & 
        \includegraphics[width=75px, height=55px]{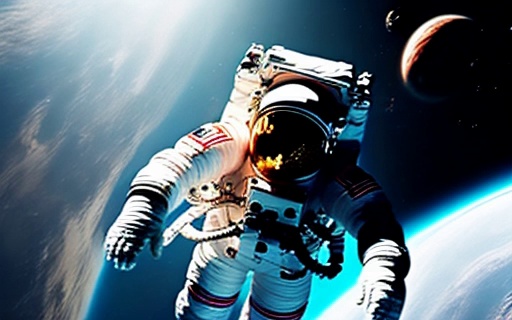} &
        \includegraphics[width=75px, height=55px]{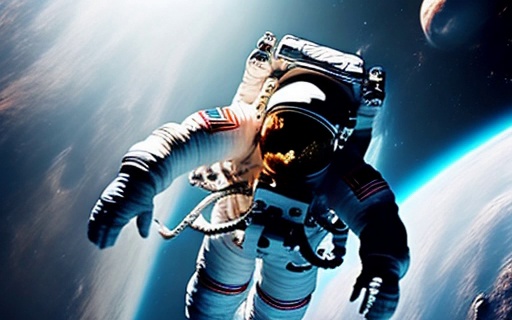} &
        \includegraphics[width=75px, height=55px]{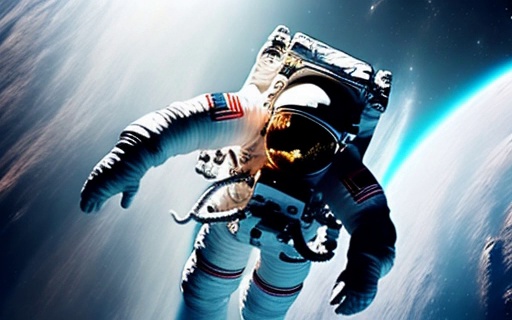} \\ 
        
        \multicolumn{3}{c}{\small \textsf{``A boat sailing along the Seine River''}} &
        \multicolumn{3}{c}{\small \textsf{``An astronaut flying in space''}} \\
        
        \includegraphics[width=75px, height=55px]{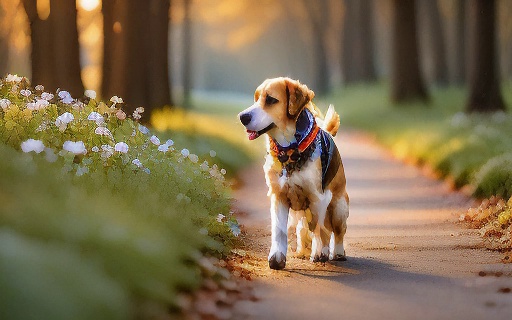} &
        \includegraphics[width=75px, height=55px]{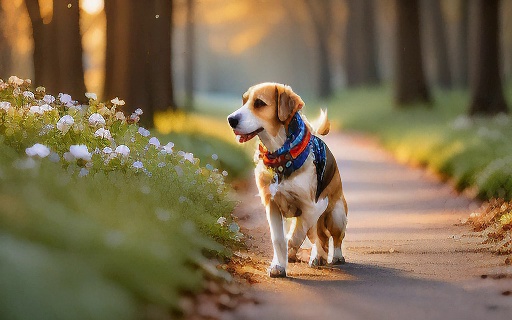} &        
        \includegraphics[width=75px, height=55px]{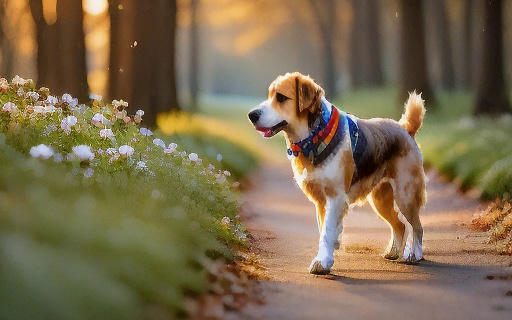} &
        \includegraphics[width=75px, height=55px]{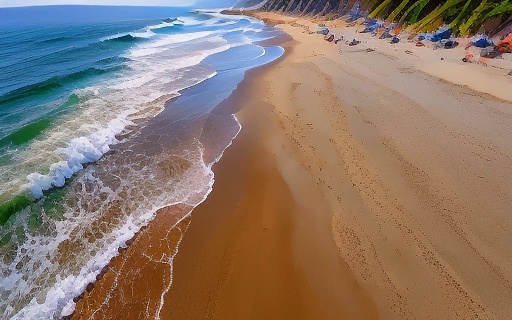} &
        \includegraphics[width=75px, height=55px]{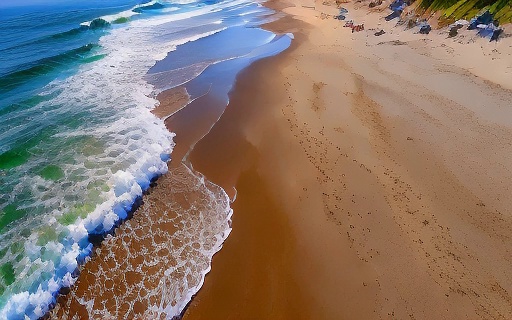} &
        \includegraphics[width=75px, height=55px]{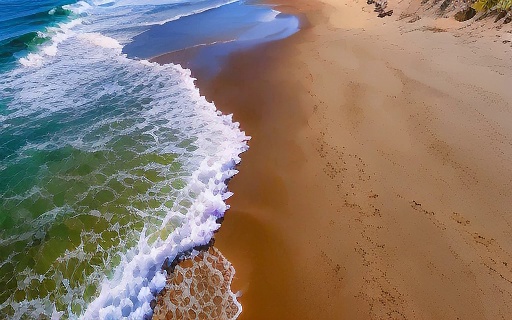} \\

        \multicolumn{3}{c}{\small \textsf{``A dog enjoying a peaceful walk''}} &
        \multicolumn{3}{c}{\small \textsf{``A beautiful coastal beach.''}} \\
    \end{tabular}
    }
    \vspace{-10pt}
    \caption{\textbf{Alignment results of \method from two different text-to-video models}, 
    including VideoCrafter2~\cite{chen2024videocrafter2} 
    (\textit{first row}), and T2V-Turbo~\cite{li2024t2vturbo} (\textit{second row}).
    More visualization results can be found in the supplementary materials. }
    \label{fig:teaser}
    \vspace{-12pt}
\end{figure*}

With the rapid advancement of computing power and the increasing scale of training data, generative diffusion models have made remarkable progress in generation quality and diversity for video generation. However, current video diffusion models often fall short of meeting user preferences in both generation quality and text-video semantic alignment, ultimately compromising user satisfaction. 

These issues often arise from the pre-training data, and filtering out all low-quality data is challenging given the vast often of pre-training data.
Specifically, two types of low-quality data are prevalent. First, regarding the videos themselves, some samples suffer from low resolution, blurriness, and temporal inconsistencies, which negatively impact the visual quality of generated videos. Second, regarding text-video pairs, mismatches between text descriptions and video content reduce the model’s ability to be controlled accurately through text prompts. Similar challenges are also seen in content generation for other modalities, such as language and image generation, where noisy pre-training data lowers output quality and reliability.

User preference alignment through Direct Preference Optimization, or DPO~\cite{rafailov2023DPO}, has been proposed and tackles these issues well for language and image generation~\cite{wallace2023diffusionDPO}.
In this paper, we focus on aligning video diffusion models with user preferences with the idea of DPO with crucial adaption modifications, termed VideoDPO, described next.

First, we introduce a comprehensive preference scoring system, \scorename, which assesses both the visual quality and semantic alignment of generated videos. We build the DPO reward model based on \scorename. While existing visual reward models~\cite{kirstain2023pick} typically focus on only one of these aspects, our experiments (see Fig.~\ref{fig:subfig_d}) show that visual quality and semantic alignment, as well as various facets of visual quality, have low correlation. Addressing a single aspect does not inherently capture the others. Thus, a comprehensive scoring system like \scorename, which integrates both dimensions, is crucial for accurate evaluation and alignment.

Second, obtaining preference annotations for generated videos is challenging due to the high cost of human labeling. To address this, we propose a pipeline that automatically generates preference pair data by strategically sampling from multiple videos conditioned on a given prompt, thereby eliminating the reliance on human annotation.

Third, to further improve the performance and efficiency of alignment training, we introduce a novel data re-weighting method, \emph{OmniScore-Based Re-Weighting}. This approach is based on the intuition that certain preference pairs, particularly those with larger quality differences, have a greater impact on alignment. By analyzing the frequency distribution, we assign higher weights to these influential samples, prioritizing them during training. Experimental results show that our method delivers significant performance improvements while producing videos with high visual fidelity and precise semantic alignment, as shown in \cref{fig:teaser}.

In summary, our contributions are:

(\textit{i}) We pioneer the adaptation of DPO to video diffusion models, addressing the unique challenges of aligning video generation outputs with user preferences.

(\textit{ii}) We introduce key adjustments to the DPO framework, including the development of \scorename, a comprehensive preference scoring system, along with an automated preference data generation pipeline and a novel re-weighting strategy to enhance alignment training efficiency.

(\textit{iii}) We validate our framework through extensive experiments conducted on three state-of-the-art open-source text-to-video models, evaluating performance across multiple metrics. The results demonstrate the robustness and effectiveness of our approach in improving both visual quality and semantic alignment.

\begin{figure*}[htbp]
    \centering
    \includegraphics[width=0.99\textwidth]{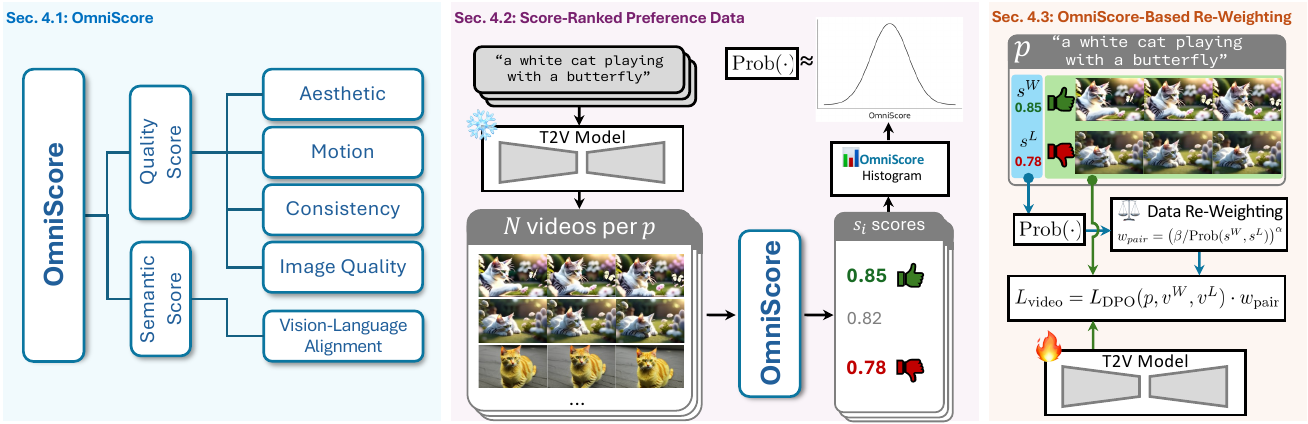} %
    \vspace{-10pt}
    \caption{\textbf{VideoDPO pipeline.} 
    We propose \scorename to rate video sample quality with multi-dimensional scores (\textit{left}). 
    For each prompt, we generate $N$ videos for each prompt $p$ and score them using \scorename. 
    The highest and lowest scores, $s^W$ and $s^L$ for the corresponding videos $v^W$ and $v^L$, form a preference pair to build the preference dataset. 
    Additionally, we compute the frequency histogram of all videos' \scorename (\textit{middle}). 
    During training, preference pairs are re-weighted based on the frequency histogram. 
    Typically, distinctive pairs which generally have lower sampling probabilities are assigned higher weights to help the model focus more on learning from them. (\textit{right}).
    }
    \label{fig:workflow}
    \vspace{-10pt}
\end{figure*}

\section{Related Work}\label{sec:relatedwork}
\subsection{Text-to-Video Diffusion Models}
The Text-to-Video (T2V) task aims to produce visually appealing videos that align with text input, ultimately striving to meet user requirements. It has wide applications across various domains, including story animation~\cite{he-animate-a-story}, controllable video generation~\cite{he-follow-your-pose,he-follow-your-click}, video game development~\cite{che2024gamegen}, and embodied artificial intelligence~\cite{chen2024igor}. The predominant approaches for video generation~\cite{SVD,wang2023modelscope,he2022lvdm,Gen2} employ diffusion-based models~\cite{sohldickstein2015deep,DDPM}. Non-diffusion frameworks~\cite{dai2023emu,wang2024emu3,park2021nerfies} have also shown significant progress. For instance, VideoCrafter~\cite{chen2023videocrafter1,chen2024videocrafter2} utilizes a 1.4 billion parameter U-Net architecture for video generation, while models~\cite{opensora,hong2022cogvideo,yang2024cogvideox} such as Open-Sora~\cite{opensora} and CogVideoX~\cite{hong2022cogvideo,yang2024cogvideox} are based on a Diffusion Transformer (DiT) backbone~\cite{Peebles2022DiT, chen2023pixartalpha}.

Given the complexity of video data, transferring diffusion pipelines to generate high-quality video content is a non-trivial task. This challenge is compounded by the necessity of implementing a series of post-training methods aimed at enhancing video quality. Existing methods include parameter efficient tuning~\cite{li2024t2vturbov2,li2024t2vturbo,he-scalecrafter,he-make-cheap-scaling}, data-centric work~\cite{he2024venhancer}, and human preference alignment~\cite{prabhudesai2024VADER,2023InstructVideo} work. Despite the continuous expansion of training datasets and computational resources, the resulting video quality often falls short of user expectations.

\subsection{RLHF and RLAIF}

Reinforcement Learning from Human Feedback (RLHF) is one of the most widely used post-training methods on large language models~\cite{yuan2023rrhf, Zhao2023SLiCHFSL, xu2024contrastive} and diffusion models. RLHF contains a reward model in the training stage and reinforcement learning stage. The reward model is trained on win-lose pairs annotated by humans by predicting the preference label. Prior works use policy-gradient~\cite{schulman2017proximal} methods to align the policy model. The two-stage training pipeline is unstable and complex. %

\paragraph{Preference alignment in diffusion models.}
DPO~\cite{rafailov2023DPO} is a reward model free method that can be easily performed on diffusion models. Despite the DPO-based methods being tested on text-to-image diffusion~\cite{wallace2023diffusionDPO} models and gaining significant process, it has been rarely tested on T2V diffusion models. 
VADER~\cite{prabhudesai2024VADER} applies a reward model to refine a video diffusion model. T2V-turbo~\cite{li2024t2vturbo,li2024t2vturbov2} exploring training consistent distillation models by reward gradients. SPO~\cite{liang2024SPO} tries to improve quality on each step of the diffusion inverse process. T2V-turbo v2~\cite{li2024t2vturbov2} also uses a reward model for refinement. 
To the best of our knowledge, we are the first to propose to apply DPO-based method on video diffusion. 

\paragraph{Visual content quality assessment.}
Previous video generation models often use metrics of Inception Score (IS)~\cite{salimans2016inceptionscore}, Fréchet inception distance (FID)~\cite{heusel2017fid}, Fréchet Video Distance (FVD)~\cite{unterthiner2018fvd}, and CLIPSIM~\cite{radford2021clip} for evaluation.
For text-to-image (T2I) models, several benchmarks~\cite{huang2023t2icompbench,wang2022imagenedit,saharia2022imagen,lee2023holistic}.
Several benchmarks~\cite{huang2023vbench, liu2023evalcrafter, liu2023fetv, t2vscore} have been proposed to comprehensively evaluate the capabilities of video generation models. These benchmarks typically assess generation quality by using pre-trained score models~\cite{radford2021clip,aesthetic,teed2020raft,Ke2021MUSIQ} to evaluate videos generated from a curated set of human-designed prompts. Other benchmarks, such as those for compositional video generation~\cite{sun2024t2vCompBench}, story generation~\cite{bugliarello-etal-2023-storybench}, chronological generation~\cite{yuan2024chronomagic}, and dynamic and motion quality~\cite{liu2024frMotionbench, liao2024dynamicbench}, focus on evaluating specific sub-tasks within video generation.

\section{Preliminaries}
\subsection{Diffusion Models}
For diffusion models, visual contents are generated by transforming a initial noise to the desired sample through multiple sequantial steps. 
It is a Markov chain process where the model continually denoises the initial noise vector \( \mathbf{x}_T \) and finally generates a sample \( \mathbf{x}_0 \). 

The generation step from $\mathbf{x}_t$ to $\mathbf{x}_{t-1}$ is given by
\[
\mathbf{x}_t \sim q(\mathbf{x}_t | \mathbf{x}_{t-1}) = \mathcal{N}(\mathbf{x}_t; \sqrt{\alpha_t} \, \mathbf{x}_{t-1}, \beta_t \mathbf{I}),
\]
where \( \beta_t \) is the variance schedule, determining the amount of noise added at each timestep \( t \).
\( \alpha_t \) is a parameter obtained by \( \alpha_t = 1 - \beta_t \) which represents the proportion of the original data retained.

The denoising model \( \epsilon_\theta \), which learns to predict the noise added to \( \mathbf{x}_0 \) for timestep $t$, is trained 
by minimizing the loss between the ground-truth \( \epsilon \) and prediction. The loss function is defined as
\[
L_{d}(\theta) = \mathbb{E}_{t, \mathbf{x}_0, \epsilon} \left[ \left\| \epsilon - \epsilon_\theta \left( \sqrt{\bar{\alpha}_t} \, \mathbf{x}_0 + \sqrt{1 - \bar{\alpha}_t} \, \epsilon, t \right) \right\|^2 \right],
\]
where \( \epsilon \) is the noise added in the forward process, and \( \bar{\alpha}_t \) is the cumulative product of \( \alpha_t \) up to timestep \( t \). 

\subsection{Direct Preference Optimization}
\label{pre:dpo}
DPO~\cite{rafailov2023DPO} is a technique used to align generative models with human preferences. 
Training on pairs of generated samples with positive and negative labels, 
the model learns to generate positive samples with higher probability and negative samples with lower probability.
DiffusionDPO~\cite{wallace2023diffusionDPO} adapts DPO for text-to-image diffusion models. 
The loss function provided in the \cite{wallace2023diffusionDPO} is defined as:
\[
L_{\text{DPO}}(x^W, x^L, c) = L(x^W, p) - L(x^L, p),
\]
where \( x^W \) and \( x^L \) represent positive and negative samples, respectively. \( L(x^W, p) \) and \( L(x^L, p) \) are losses for positive and negative parts, encouraging the model to generate samples closer to preferences.

\section{VideoDPO}

\subsection{\scorename}
The quality of generated videos is influenced by multiple factors, which can be grouped into two main categories: visual quality and semantic alignment. %
Visual quality includes the clarity and richness of detail within each frame, \ie, intra-frame quality, and the smoothness and coherence between frames, \ie, inter-frame motion and consistency. %
Semantic alignment, on the other hand, 
focuses on whether the generated video accurately follows the text prompt.
Inspired by VBench~\cite{huang2023vbench}, we propose a scoring approach for video generation, 
\scorename, which comprehensively accounts for both visual quality and semantic alignment 
of generated videos.
\scorename incorporates both quality and semantic sub-scores, 
specifically designed to evaluate video generation on three primary dimensions: 
the fidelity and aesthetics of visual quality, the smoothness of inter-frame transitions, 
and the level of semantic alignment with the text. 
Each model for these dimensions is provided in the Appendix. 
This holistic approach enables a balanced method for preference pair data generation.

\paragraph{Intra-frame quality. } 
Intra-frame quality includes two main metrics, image quality and aesthetic appeal. 
These metrics assess the visual quality of individual frames measuring image fidelity and aesthetic attractiveness. 
They provide a thorough evaluation of the frame-level visual detail, 
ensuring each frame is not only of high-fidelity but also visually engaging.

\paragraph{Inter-frame quality. } 

Inter-frame quality focuses on the relationships between consecutive frames, 
examining how well they connect over time. 
This dimension includes metrics for subject consistency and background consistency, 
which assesses the stability of key elements across frames, 
ensuring that the main subject and background remain visually coherent. 
Additionally, it evaluates motion dynamics through three metrics: temporal flickering, 
motion smoothness, and the degree of motion dynamics. 
These collectively examine the video’s fluidity, 
ensuring smooth transitions between frames, minimizing visual disruptions, 
and maintaining a natural level of movement.
By considering these aspects, we aim to ensure that the video maintains visual continuity 
and avoids disruptions that can detract from the viewing experience.

\paragraph{Text-video semantic alignment. } 
Semantic alignment evaluates how closely the video content aligns with the text prompt. Using a foundational vision-language model, this score measures how accurately the video reflects the text and captures the user’s intent. %

\subsection{Score-Ranked Preference Data Generation}
\label{method:dataset}

To construct the dataset of preference pairs, the scoring method, \scorename, 
is employed in combination with a best \vs worst selection strategy. 
For each prompt, our system generates multiple videos and a preference pair is selected. 
Specifically, the video with the highest \scorename is identified 
as the preferred video $v^W$, while the video with the lowest score is designated as the negative one $v^L$, as shown in Fig.~\ref{fig:workflow}. 
We utilize VidProm~\cite{wang2024vidprom}, a dataset of human-written text-to-video prompts, in our data construction process, enabling the model to better adapt to the distribution of real-world human inputs.

\paragraph{Video generation and scoring.}

Given a text prompt \( p \), we generate a set of \( N \) videos \( \{v_1, v_2, \dots, v_N\} \), 
using the pre-trained video generation model that we aim to align. 
For each generated video \( v_i \), 
we apply the \scorename  model \( S \) to evaluate its quality 
conditioned on the text prompt \( p \). 
This scoring model assigns a score to each video:
\begin{equation}
s_i = S(v_i, p), \, \text{for} \; i=1, 2, \dots, N\,.
\end{equation}
Here, \( s_i \) denotes the \scorename assigned to video \( v_i \) 
given its prompt \( p \). 
This scoring step creates a quantitative basis for comparing videos generated from the same prompt.

\paragraph{Preference pair selection. }

We select preference pairs \( (v_i, v_j) \) from the \( N \) generated videos according to their \scorename \( \{s_1, s_2, \dots, s_N\} \).
We select the video with the highest score 
as the winning sample \( v^W \) and the video with the lowest score as 
the negative sample \( v^L \). This selection process is formalized as follows:
\begin{equation}
    (v^W, v^L) = (v_i, v_j),\, i = \arg\max_{i} s_i,\, j = \arg\min_{j} s_j\,.
\end{equation}

By constructing preference pairs consistently with maximally contrasting scores, we aim to establish clear distinctions between the
preferred, or winning, and the less-preferred, or losing video samples. This strategy serves as a strong foundation for training the alignment model. We discuss several other selection strategies in \cref{exp:pair}. %

\subsection{\scorename-Based Data Re-Weighting}

Previous DPO training directly uses winning and losing preference pairs, for example, those generated as described in \cref{method:dataset}. %
However, we find the score difference between some winning 
and negative samples can be minimal, or in some cases, nearly identical, making it challenging for the model to effectively distinguish these samples with minor differences. 
To address this, we propose assigning higher weights to preference pairs with clearer distinctions, 
enabling the model to focus on those pairs that could provide more meaningful alignment cues. 
Our approach significantly enhances the model's ability to learn meaningful alignment preferences. %

Specifically, we first construct a histogram of \scorename $s$ of each generated video, %
including \(K\)\,\(\times\)\,\(N\) videos from  \( K \) 
prompts in total. %
We denote \( p(\cdot) \) as the frequency 
and we define a function \( p(\cdot) \) to approximate the probability of a video based on its frequency within these bins.

For each winning-losing pair, we define the pair probability as 
the geometric mean of their individual probabilities, 
i.e., \( \text{prob}(s^W, s^L) = \sqrt{p(s^W) \cdot p(s^L)} \), 
where \( s^p \) and \( s^n \) represent the scores of the winning (positive) and losing (negative) samples, 
respectively. We define the re-weighting factor for each pair as:
\begin{equation}
w_{\text{pair}} = \left( \beta / \text{prob}(s^W, s^L) \right)^{\alpha}\,.
\end{equation}
Here, \( \beta \) is a constant set to the approximate probability of the most frequent sample, 
and \( \alpha \) is a tuning hyperparameter. When \( \alpha \) equals to \( 0 \), no re-weighting is applied, 
and all pairs have equal weight. A larger \( \alpha \) increases the weight for pairs with lower probability.

The final training loss for each video pair is defined as:
\begin{equation}
L_{\text{video}} = L_{\text{DPO}}(p, v^W, v^L) \cdot w_{\text{pair}}\,,
\end{equation}
where \( L_{\text{DPO}} \) refers to the DPO loss described in \cref{pre:dpo}. 
The re-weighting factor \( w_{\text{pair}} \) adjusts the 
impact of each pair, encouraging the model to learn more effectively from those with clearer distinctions.

\section{Experiment}
\subsection{Experiment Setup}
\paragraph{Baselines.}
We compare our pipeline with several state-of-the-art open-source models 
for text-to-video generation: VideoCrafter-v2(VC2)~\cite{chen2024videocrafter2}, T2V-Turbo(Turbo)~\cite{li2024t2vturbo}, and CogVideo~\cite{hong2022cogvideo}. 
These models are utilized as baselines in our alignment experiments. 
Additionally, we include VADER\cite{prabhudesai2024VADER}, which directly fine-tunes 
video diffusion models in several final steps using the differentiable reward model. 
We compare our method with their publicly released weights.

\paragraph{Metrics. }
To evaluate our method and the baselines, 
we use the following metrics: 
VBench, a widely recognized benchmark that assesses both 
quality and semantic alignment in video generation across 16 hierarchical dimensions, 
providing fine-grained evaluation. HPS (V)~\cite{wu2023human} and PickScore~\cite{kirstain2023pick} are also included as metrics;
both are trained on large-scale human preference datasets and are designed to 
predict scores of human preference for generated videos.

\paragraph{Implementation details. }
We train the video diffusion models for 3000 steps with a global batch size of 8, 
using the AdamW optimizer with a learning rate of 6e-6. 
During training, the re-weighting algorithm hyper-parameters are set to $\alpha=0.72$ and $\beta=1$. 
$K=10,000$ human-written prompts from VidProm~\cite{wang2024vidprom} are used for alignment training. 
For each prompt, the number of generated videos $N$ is set to 4. 
The bin width for the distribution of video \scorename scores is set to 0.01.
All experiments are conducted on 4 Nvidia A100 GPUs.

\begin{table}[ht]
    \small 
    \setlength{\tabcolsep}{4pt}
    \centering
        \resizebox{0.48\textwidth}{!}{
        \begin{tabular}{clccccc}
            \toprule
            & \multirow{2}{*}{\textbf{Model}} & \multicolumn{3}{c}{\textbf{VBench (\%)}} & \multirow{2}{*}{\textbf{HPS (V)}} & \multirow{2}{*}{\textbf{PickScore}} \\
            \cmidrule(lr){3-5}
            & & \textbf{Total} & \textbf{Quality} & \textbf{Semantics} &  & \\
            \midrule
            \multirow{4}{*}{\rotatebox{90}{VC2}} 
            & Baseline\cite{chen2024videocrafter2} & 80.44 & 82.20 & 73.42 & 0.258 & 20.65 \\
            & SFT & 78.78 & 79.90 & 74.32 & 0.258 & 20.35 \\
            & VADAR\cite{prabhudesai2024VADER} & 80.59 & 82.46 & 73.09 & 0.259 & 20.62\\
            & \GR VideoDPO & \GR \textbf{81.93} & \GR 83.07 & \GR 77.38 & \GR {0.261} & \GR {20.65} \\
            \midrule
            \multirow{2}{*}{\rotatebox{90}{Turbo}} 
            & Baseline\cite{li2024t2vturbo} & 80.95 & 82.71 & {73.93} & 0.262 & 21.15\\
            & \GR VideoDPO & \GR \textbf{81.80} & \GR {83.80} & \GR 73.81 & \GR 0.260 & \GR {21.18} \\
            \midrule
            \multirow{3}{*}{\rotatebox{90}{CogVid.}} 
            & Baseline~\cite{hong2022cogvideo} & 79.30 & 82.35 & 67.10 & -& {19.81}\\
            & SFT & 79.64 & 82.74 & {67.23} & - & 19.79\\
            & \GR VideoDPO & \GR \textbf{79.80} & \GR {83.00} & \GR 66.99 & \GR - & \GR 19.79\\
            \bottomrule
        \end{tabular}
        }
    \vspace{-5px}
    \caption{
        \textbf{VideoDPO alignment performance}. We apply our proposed \method on three state-of-the-art open-source models and evaluate performance on VBench, 
        HPS (V), and PickScore. After training with \method, 
        all models achieve the best performance on VBench, 
        with improvements also observed on HPS (V) or PickScore, demonstrating the effectiveness of our approach. 
        }\label{tab:quan}
    \label{tab:main}
    \vspace{-10pt}
\end{table}

\subsection{Dataset Analysis}
\paragraph{Score distribution and sub-dimension correlations.}

\begin{figure*}[ht]
    \centering
    \begin{subfigure}[b]{0.24\linewidth}
        \centering
        \includegraphics[width=1\textwidth]{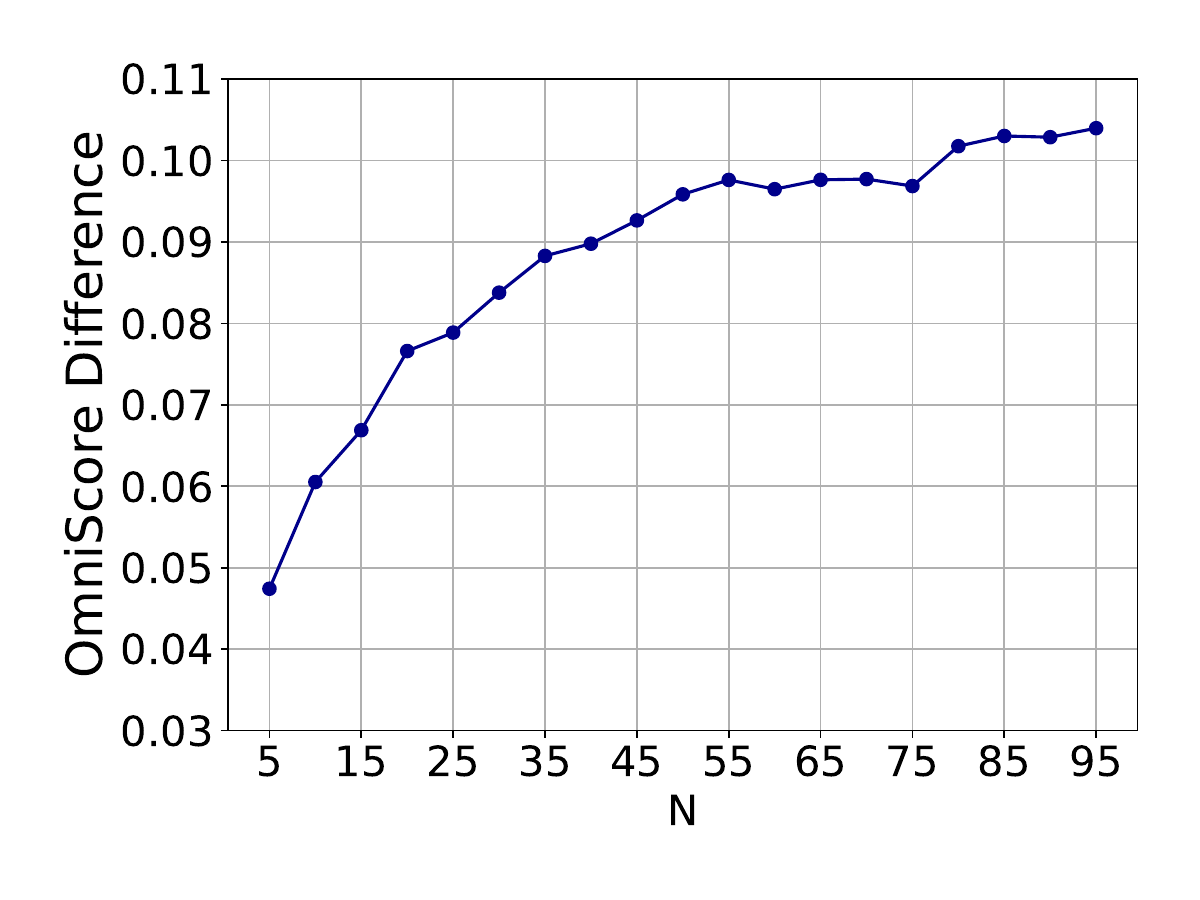}
        \vspace{-16px}
        \caption{}
        \label{fig:subfig_a}
    \end{subfigure}
    \hspace{-8px}
    \begin{subfigure}[b]{0.24\linewidth}
        \centering
        \includegraphics[width=1\textwidth]{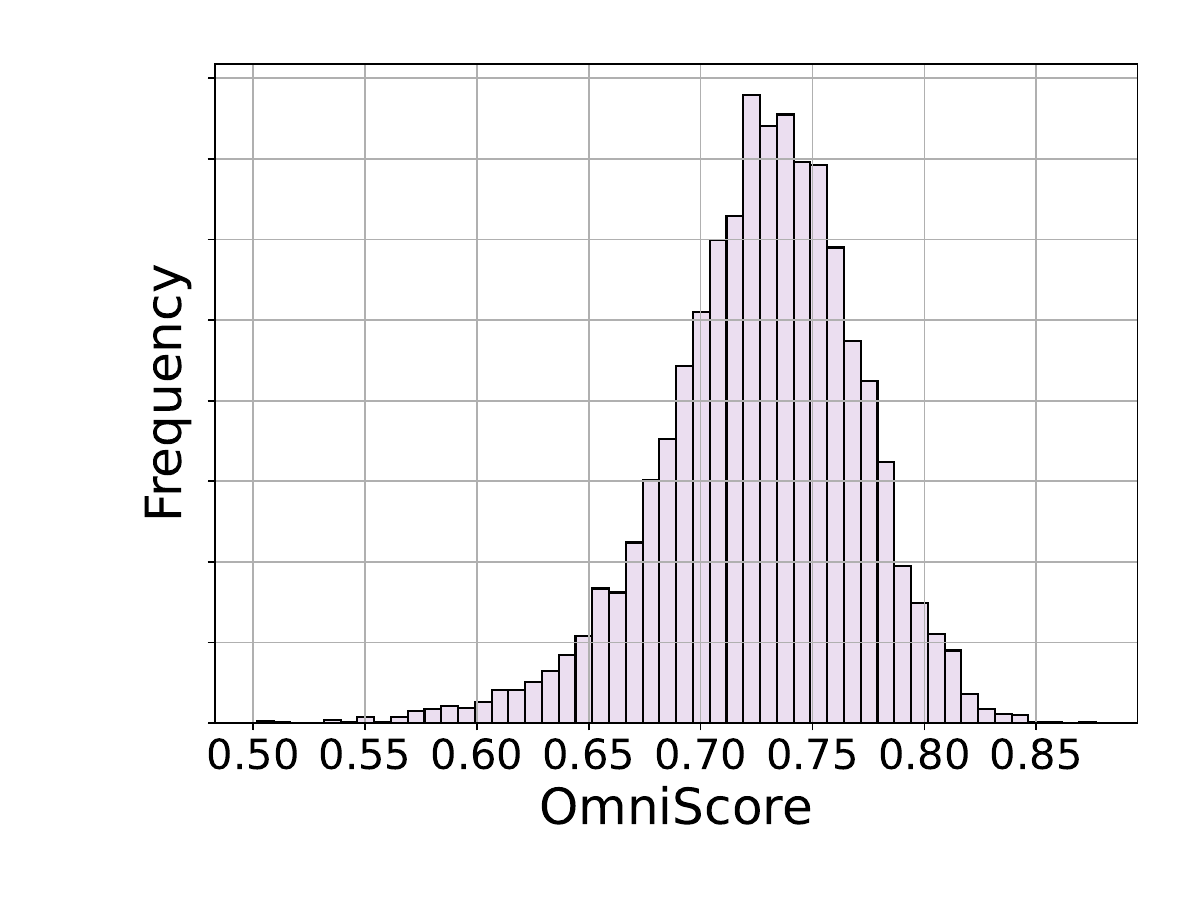}
        \vspace{-16px}
        \caption{}
        \label{fig:subfig_b}
    \end{subfigure}
    \hspace{-8px}
    \begin{subfigure}[b]{0.24\linewidth}
        \centering
        \includegraphics[width=1\textwidth]{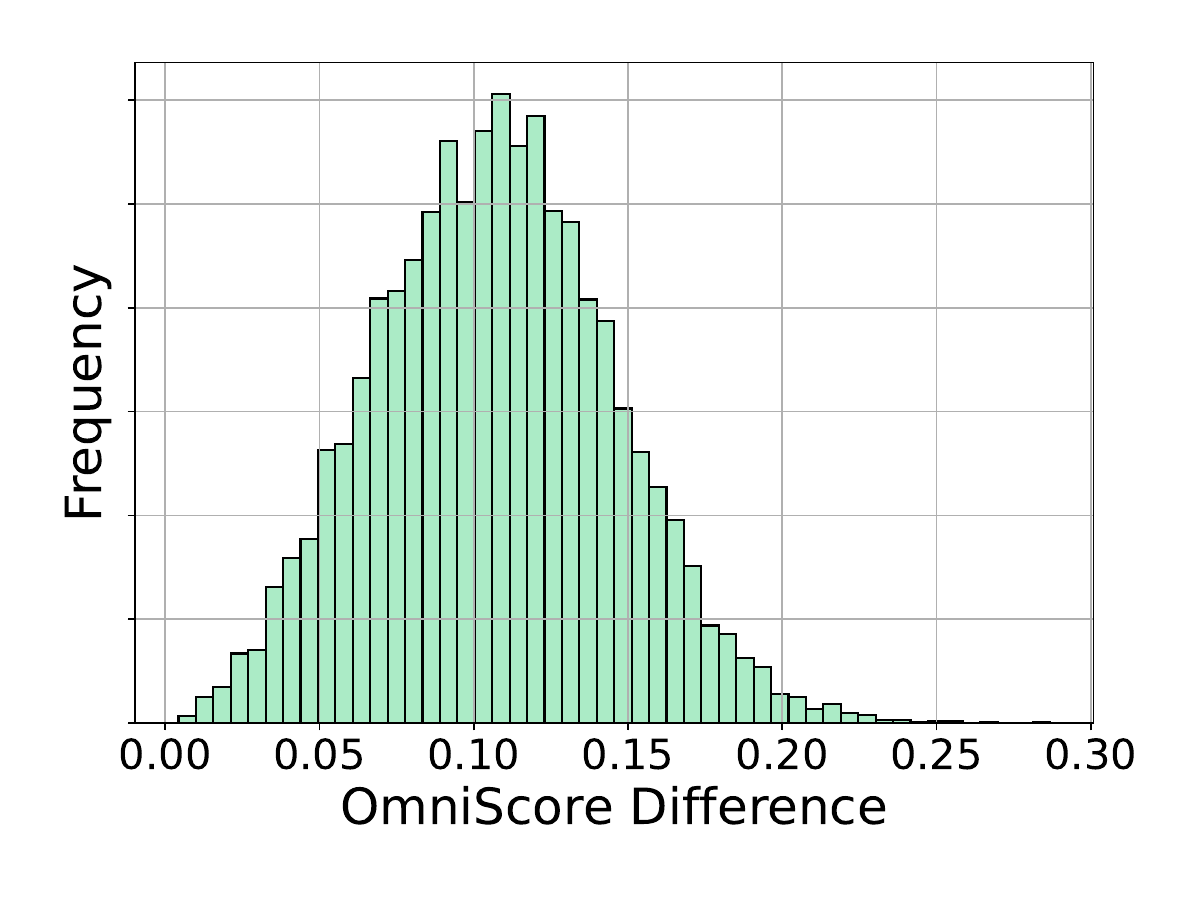}
        \vspace{-16px}
        \caption{}
        \label{fig:subfig_c}
    \end{subfigure}
    \hspace{-8px}
    \begin{subfigure}[b]{0.24\linewidth}
        \centering
        \includegraphics[width=1\textwidth]{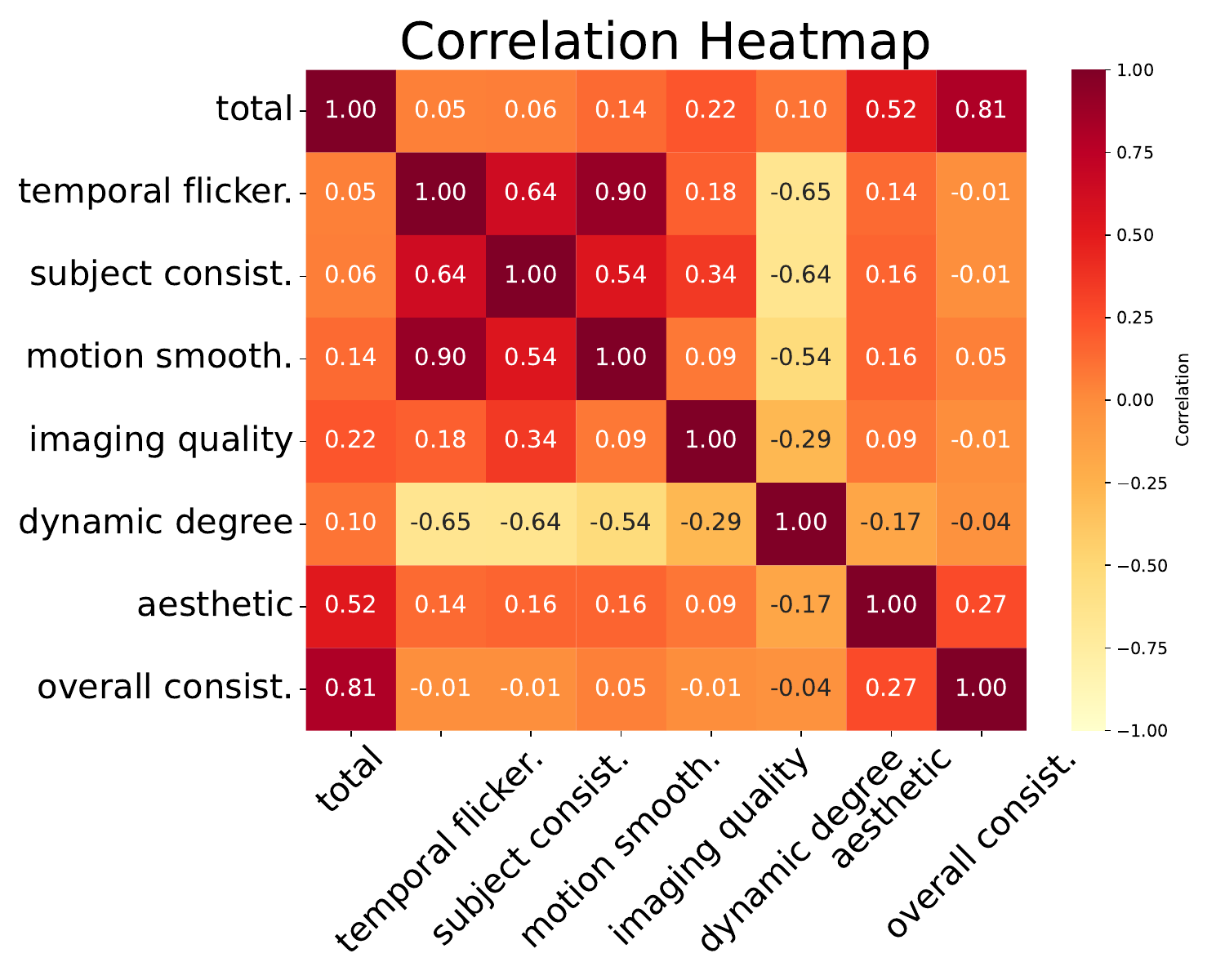}
        \vspace{-16px}
        \caption{}
        \label{fig:subfig_d}
    \end{subfigure}
    \vspace{-8px}
    \caption{
     \textbf{Analysis of \scorename.} 
     \textbf{(a)} The difference between the maximum and minimum \scorename among $N$ videos as $N$ increases. 
     \textbf{(b)} Histogram of \scorename. 
     \textbf{(c)} Histogram of the difference in \scorename between two samples in a preference pair. 
     \textbf{(d)} Correlation heatmap of the \scorename across dimensions.
    }
    \label{fig:distribution}
    \vspace{-5pt}
\end{figure*}

We analyze the dataset by examining the \scorename distribution, the score range, \ie, the difference between maximum and minimum scores, for each prompt, and the correlations among individual scoring metrics. To quantify these correlations, we calculate Pearson correlation coefficients. For each prompt, $N$ videos are generated and assigned \scorename, enabling us to assess the distribution of score differences, \ie, the range between the highest and lowest scores, within each set of $N$ videos. \cref{fig:distribution} presents both the overall score distribution and the distribution of score differences between video pairs.

\begin{table*}[ht]
    \small
    \centering
    \setlength{\tabcolsep}{3.5pt}
    \renewcommand{\arraystretch}{1.1}
    \begin{tabular}{clc|ccccccccccc}
    \toprule
    &\textbf{Model}     & \makecell{\textbf{Total}} & \makecell{\textbf{Motion} \\ \textbf{smooth.}} & \makecell{\textbf{Dynamic} \\ \textbf{degree}} & \makecell{\textbf{Aesthetic} \\ \textbf{quality}} & \makecell{\textbf{Object} \\ \textbf{class}} & \makecell{\textbf{Multiple} \\ \textbf{objects}} & \makecell{\textbf{Human} \\ \textbf{action}} & \makecell{\textbf{Spatial} \\ \textbf{relation.}} & \textbf{Scene} & \makecell{\textbf{Appear.} \\ \textbf{style}} & \makecell{\textbf{Subject} \\ \textbf{consist.}} & \makecell{\textbf{Back.} \\ \textbf{consist.}} \\
    \midrule
    \multirow{2}{*}{\rotatebox{90}{VC2}} &Baseline~\cite{chen2024videocrafter2}             & 80.44 & 97.73 & 42.50 & 63.13 & 92.55 & 40.66 & 95.00 & 35.86 & 55.29 & 87.84 & 96.85 & 98.22 \\
    &  \GR VideoDPO   & \GR 81.93 & \GR 92.18 & \GR 32.64 & \GR 63.18 & \GR 97.15 & \GR 52.29 & \GR 99.00 & \GR 48.71  & \GR 71.07 & \GR 88.65 & \GR 95.69 & \GR 96.98 \\
    \midrule
    \multirow{2}{*}{\rotatebox{90}{Turbo}} &Baseline~\cite{li2024t2vturbo}           & 80.95 & 87.27 & 27.78 & 68.57 & 93.20 & 51.83 & 96.00 & 40.98 & 62.67 & 85.07 & 96.12 & 97.62 \\
    & \GR VideoDPO       & \GR 81.80  & \GR 88.85 & \GR 29.86 & \GR 68.98 & \GR 93.59 & \GR 51.98 & \GR 94.00 & \GR 37.68 & \GR 65.23 & \GR 86.05 & \GR 96.10 & \GR 97.68\\
    \midrule
    \multirow{2}{*}{\rotatebox{90}{CogV.}} 
    &Baseline~\cite{hong2022cogvideo}       & 79.30 & 89.64 & 31.25 & 61.25 & 80.06 & 52.67 & 85.00 & 55.19 & 44.10 & 80.60 & 95.58 & 97.56 \\
    & \GR VideoDPO           & \GR {79.80} & \GR 88.64 & \GR 38.89 & \GR 58.64 & \GR 77.22 & \GR 54.04 & \GR 81.00 & \GR 54.90 & \GR 45.69 & \GR 79.73 & \GR 94.67 & \GR 96.64 \\
    \bottomrule
    \end{tabular}
    \vspace{-5pt}
    \caption{
        \textbf{Comparison of sub-dimension scores before and after alignment} on VBench for VC2, T2VTurbo and CogVideo.
        }
    \label{tab:vben}
    \vspace{-5pt}
\end{table*}

\subsection{Aligning Video Diffusion Models}
In this section, we evaluate our approach through 
both quantitative and qualitative results by testing on various 
text-to-video models. For quantitative evaluation, we utilize VBench, HPS (V), 
and PickScore, covering both non-human and human preference metrics.
To evaluate semantic alignment and visual quality, we analyze intra-frame aspects, 
examining image fidelity and aesthetic appeal to ensure each frame aligns well with the prompt. 
Additionally, for inter-frame analysis, we assess temporal consistency, 
focusing on whether the background and main foreground objects remain coherent across frames. 

\paragraph{Quantitative results. }
We present our results in \cref{tab:main}, where we evaluate state-of-the-art open-source text-to-video models, 
including VC2, T2VTurbo, and CogVideo. 
After alignment using our approach, 
all models show performance improvements, with consistent gains on the VBench metric. 
Models such as VC2 and T2VTurbo also achieve higher scores on human preference metrics, 
including HPS (V) and PickScore, demonstrating the generalizability of our approach. 
We do not report CogVideo on HPS (V) as this score appears to be insensitive to CogVideo, possibly due to the low quality generation, given its early release date.
The detailed performance results on VBench are presented in \cref{tab:vben}.
In comparison to other RLHF methods like VADAR, our approach yields superior results 
in both semantic and visual quality aspects. This improvement is attributed to 
our use of preference pairs derived from a more comprehensive feedback signal, 
both quality(intra-frame and inter-frame levels) and semantic criteria. 
However, methods like VADER can only optimize a single differentiable reward model, 
limiting improvements in other dimensions. 
Training with VADER on multiple reward models 
simultaneously will significantly increase the computational cost, making it difficult to scale.

\paragraph{Intra-frame qualitative analysis. }
\begin{figure*}[htbp]
    \centering
    \resizebox{1.0\textwidth}{!}{
    \begin{tabular}{p{2pt}c@{}c@{}cc@{}c@{}c}
        \multicolumn{4}{c}{\emph{Visual Quality}} & \multicolumn{3}{c}{\emph{Semantic Alignment}} \\
        \multirow{1}{*}[35px]{\rotatebox{90}{VC2~\cite{chen2024videocrafter2}}} & 
        \includegraphics[width=75px, height=55px]{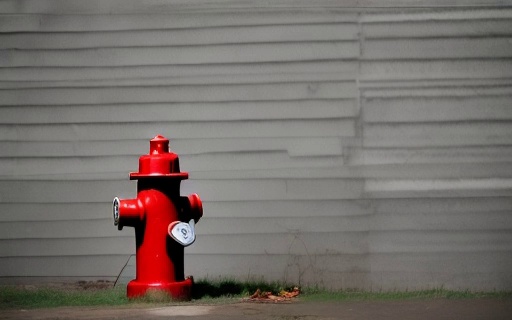} & 
        \includegraphics[width=75px, height=55px]{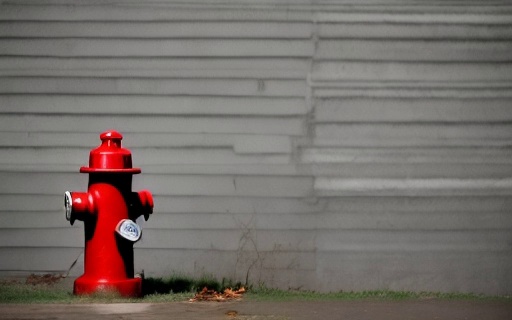} & 
        \includegraphics[width=75px, height=55px]{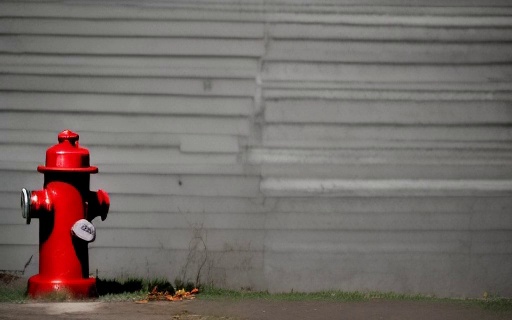} & 
        \includegraphics[width=75px, height=55px]{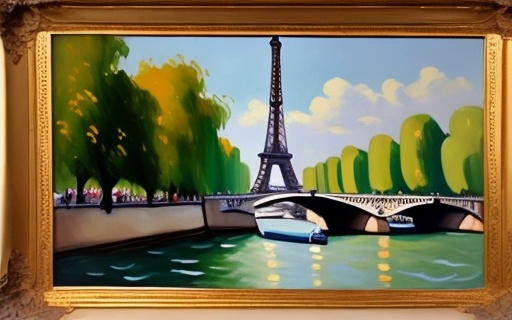} & 
        \includegraphics[width=75px, height=55px]{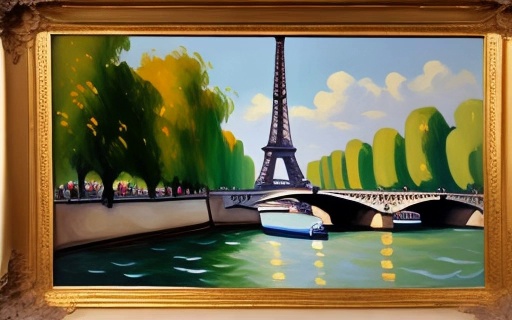} & 
        \includegraphics[width=75px, height=55px]{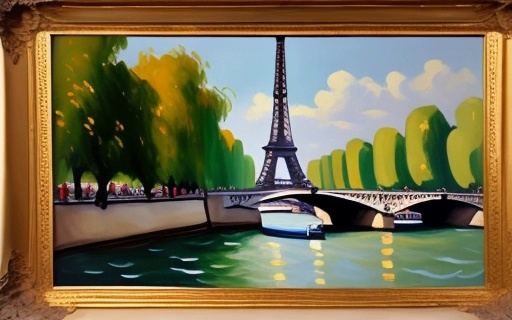} \\ 
        \multirow{1}{*}[45px]{\rotatebox{90}{VADER~\cite{prabhudesai2024VADER}}} & 
        \includegraphics[width=75px, height=55px]{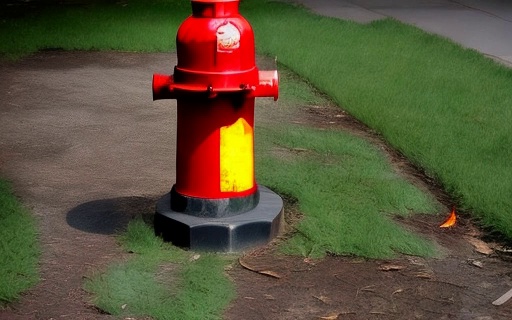} & 
        \includegraphics[width=75px, height=55px]{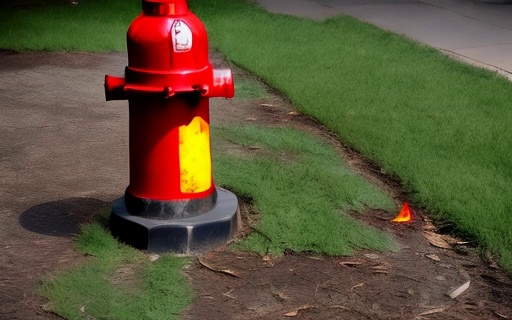} & 
        \includegraphics[width=75px, height=55px]{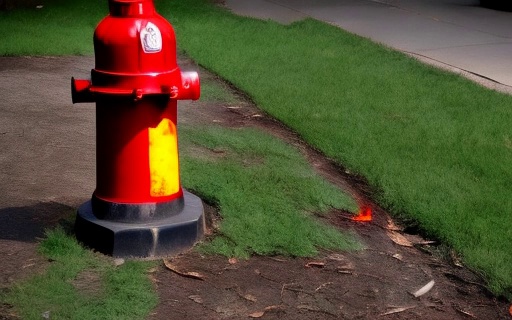} & 
        \includegraphics[width=75px, height=55px]{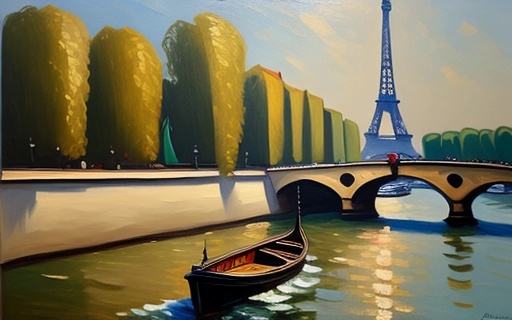} & 
        \includegraphics[width=75px, height=55px]{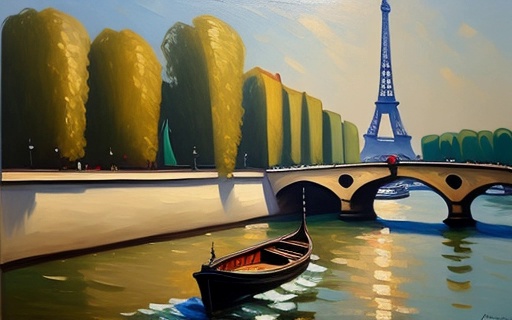} & 
        \includegraphics[width=75px, height=55px]{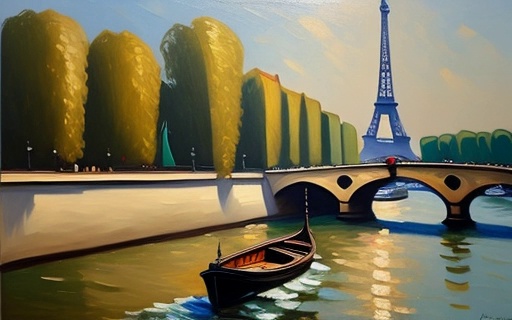}\\ 
        \multirow{1}{*}[40px]{\rotatebox{90}{\method}} & 
        \includegraphics[width=75px, height=55px]{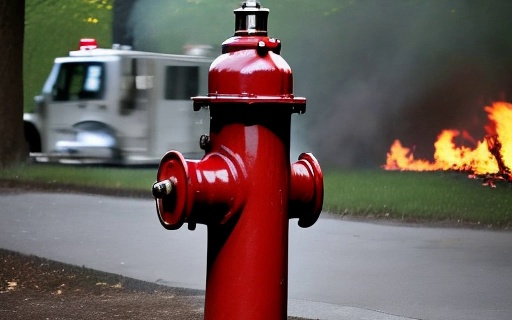} & 
        \includegraphics[width=75px, height=55px]{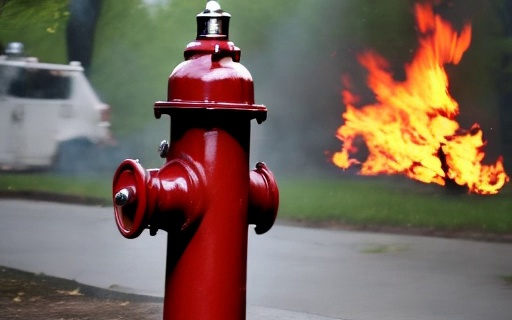} & 
        \includegraphics[width=75px, height=55px]{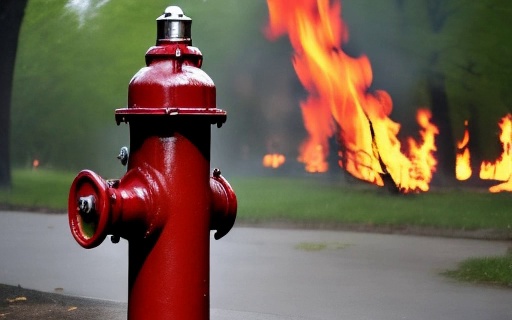} & 
        \includegraphics[width=75px, height=55px]{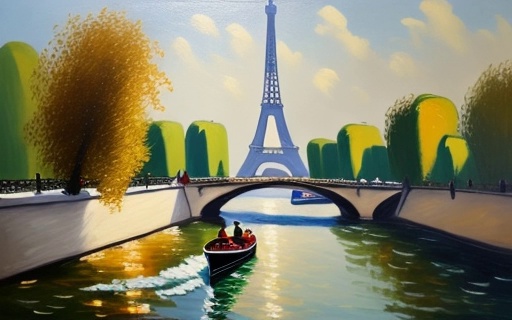} & 
        \includegraphics[width=75px, height=55px]{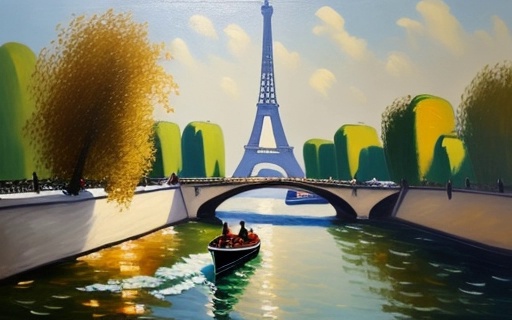} & 
        \includegraphics[width=75px, height=55px]{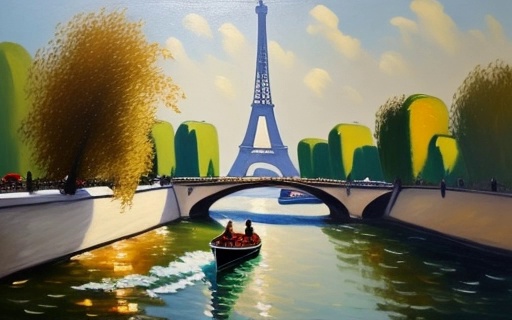} \\ 
        &
        \multicolumn{3}{c}{\small \textsf{``A fire hydrant''}} &
        \multicolumn{3}{c}{\small \textsf{``A {boat} with the Eiffel Tower in background''}} \\

        \multirow{1}{*}[40px]{\rotatebox{90}{Turbo~\cite{li2024t2vturbo}}} & 
        \includegraphics[width=75px, height=55px]{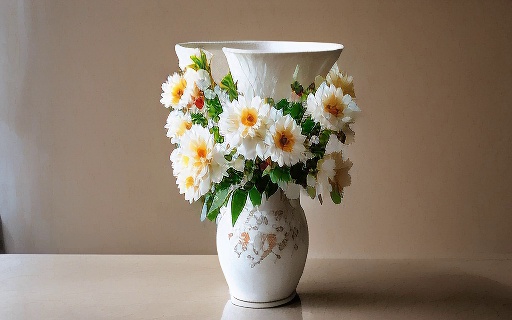} & 
        \includegraphics[width=75px, height=55px]{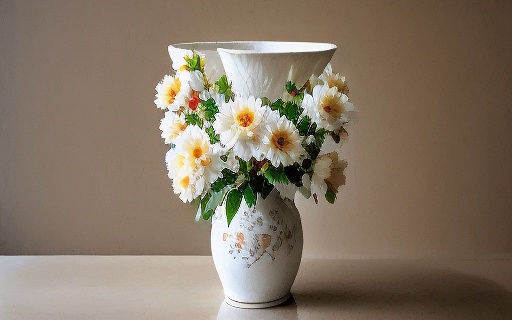} & 
        \includegraphics[width=75px, height=55px]{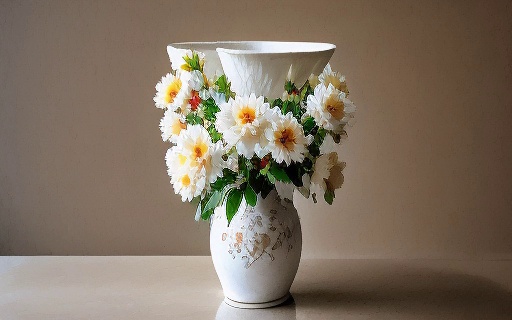} & 
        \includegraphics[width=75px, height=55px]{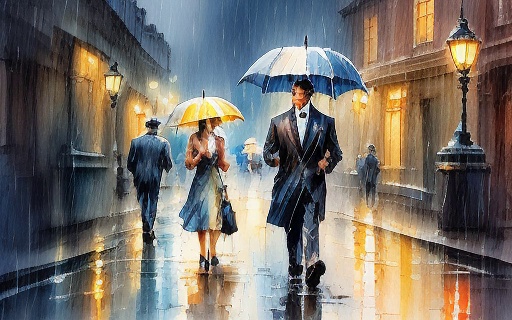} & 
        \includegraphics[width=75px, height=55px]{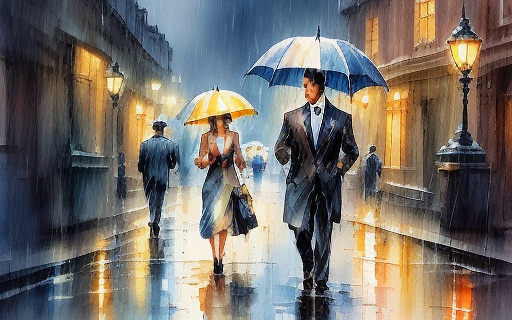} & 
        \includegraphics[width=75px, height=55px]{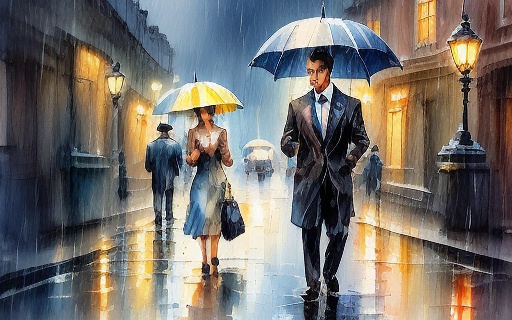}  \\ 
        \multirow{1}{*}[40px]{\rotatebox{90}{\method}} & 
        \includegraphics[width=75px, height=55px]{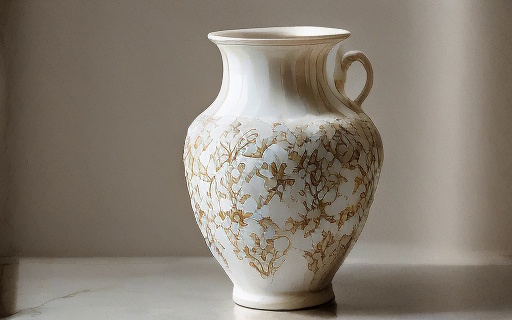} & 
        \includegraphics[width=75px, height=55px]{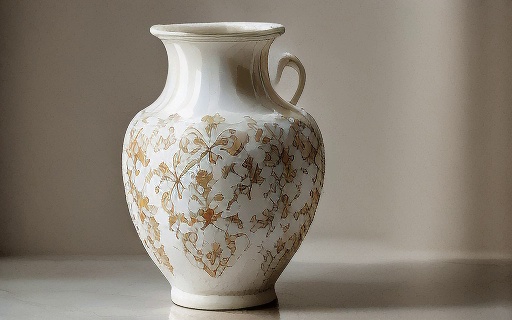} & 
        \includegraphics[width=75px, height=55px]{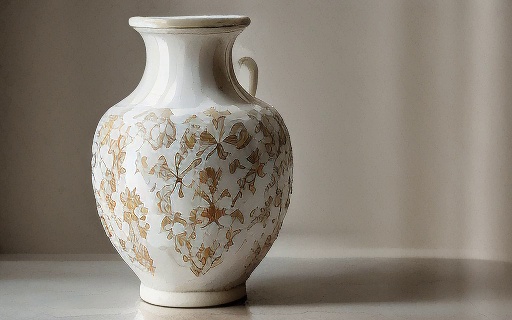} & 
        \includegraphics[width=75px, height=55px]{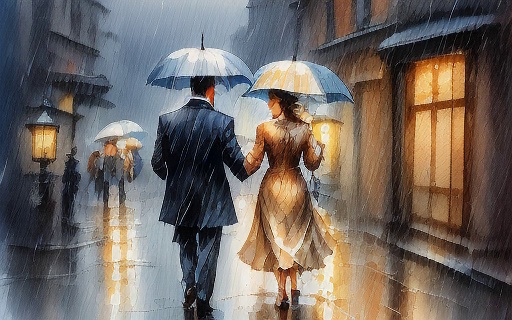} & 
        \includegraphics[width=75px, height=55px]{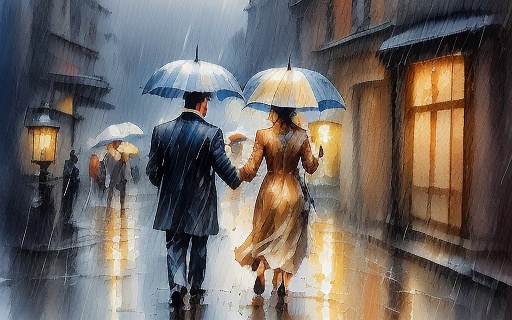} & 
        \includegraphics[width=75px, height=55px]{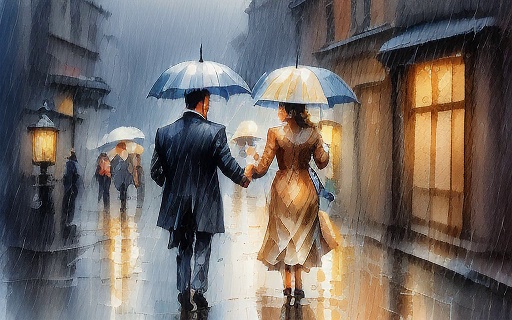} \\ 
        
        &
        \multicolumn{3}{c}{\small \textsf{``A white vase''}} &
        \multicolumn{3}{c}{\small \textsf{``A {couple} in formal evening with umbrellas''}} \\

        \multirow{1}{*}[38px]{\rotatebox{90}{CogV.~\cite{hong2022cogvideo}}} & 
        \includegraphics[width=75px, height=55px]{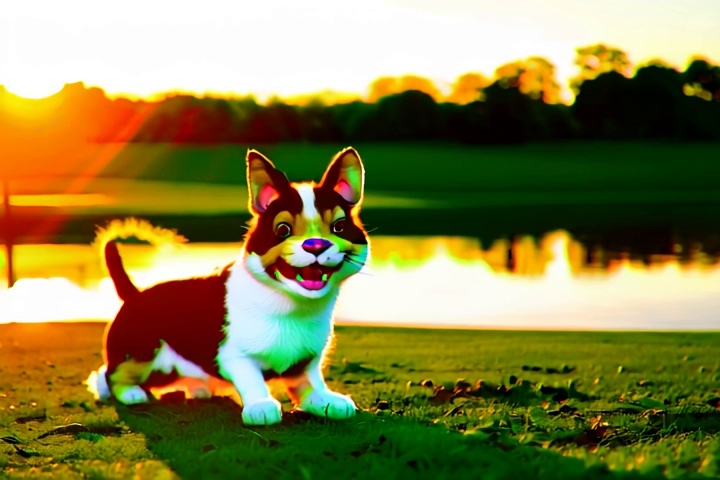} & 
        \includegraphics[width=75px, height=55px]{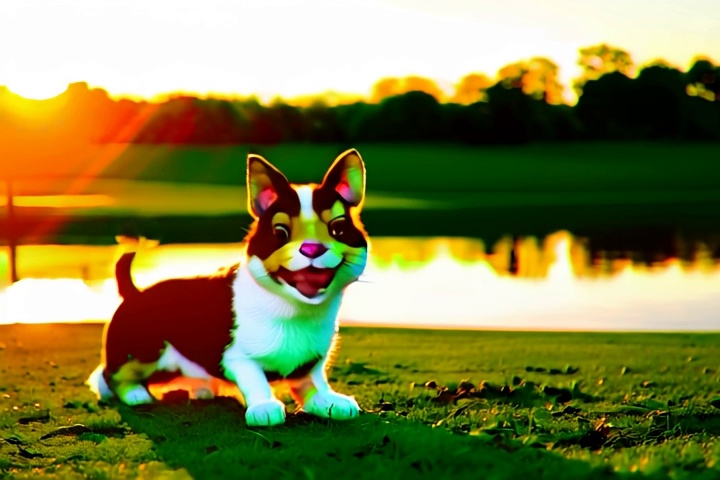} & 
        \includegraphics[width=75px, height=55px]{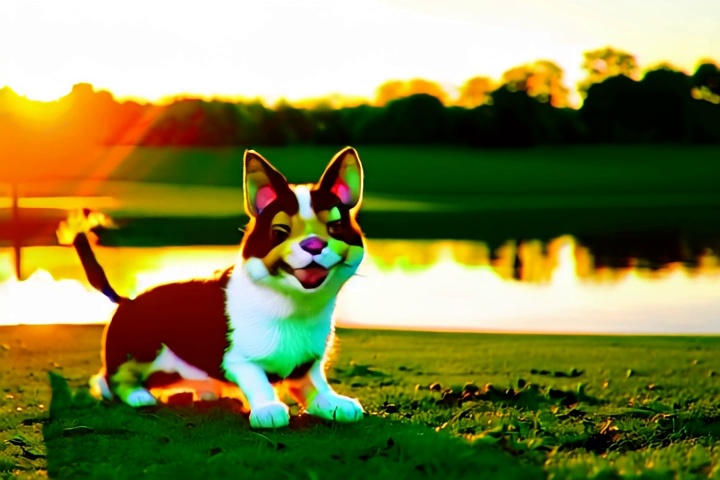} & 
        \includegraphics[width=75px, height=55px]{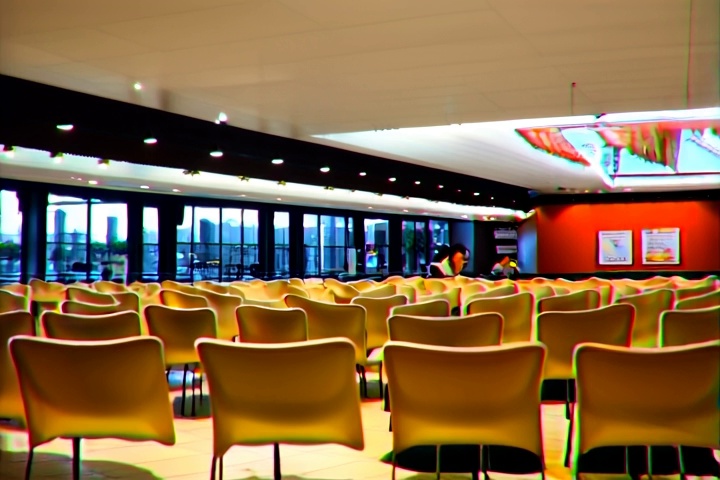} & 
        \includegraphics[width=75px, height=55px]{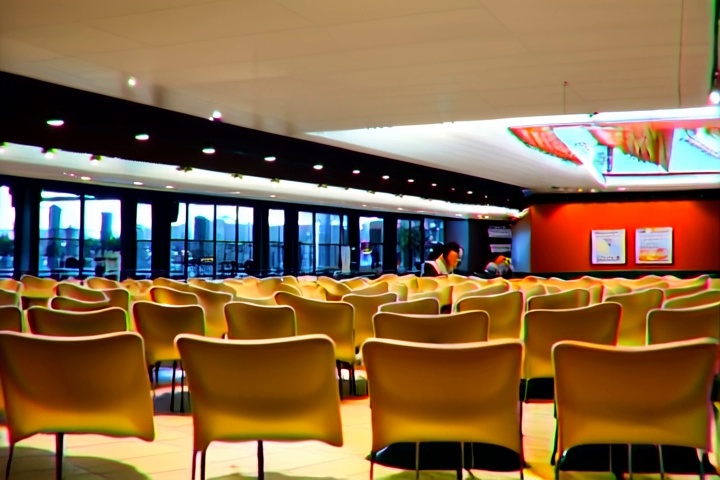} & 
        \includegraphics[width=75px, height=55px]{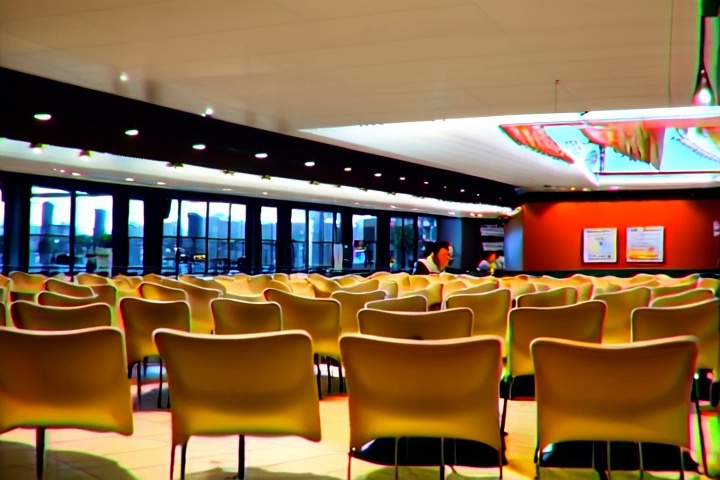} \\ 
        \multirow{1}{*}[40px]{\rotatebox{90}{\method}} & 
        \includegraphics[width=75px, height=55px]{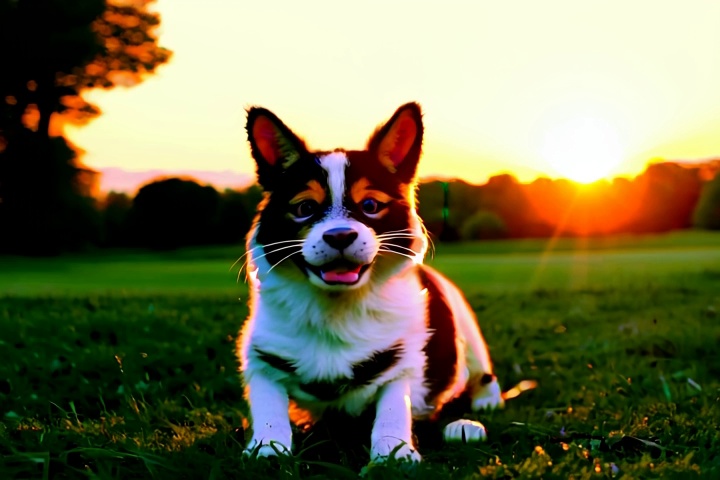} & 
        \includegraphics[width=75px, height=55px]{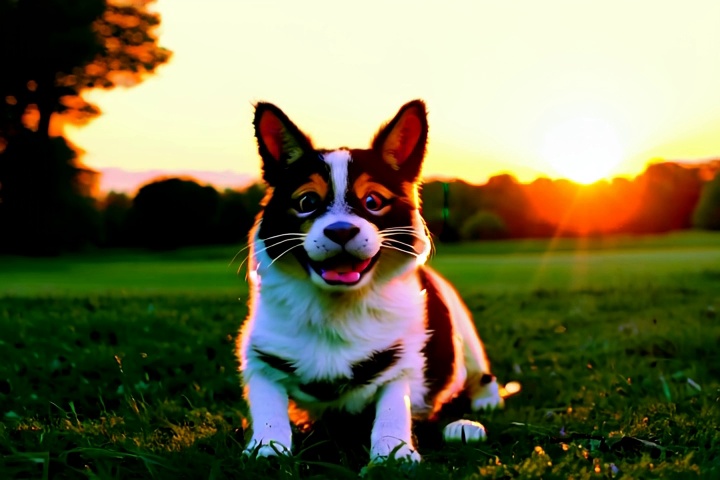} & 
        \includegraphics[width=75px, height=55px]{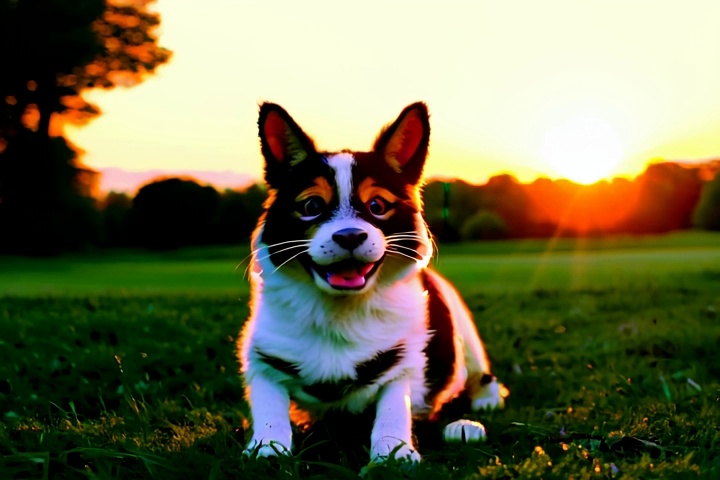} & 
        \includegraphics[width=75px, height=55px]{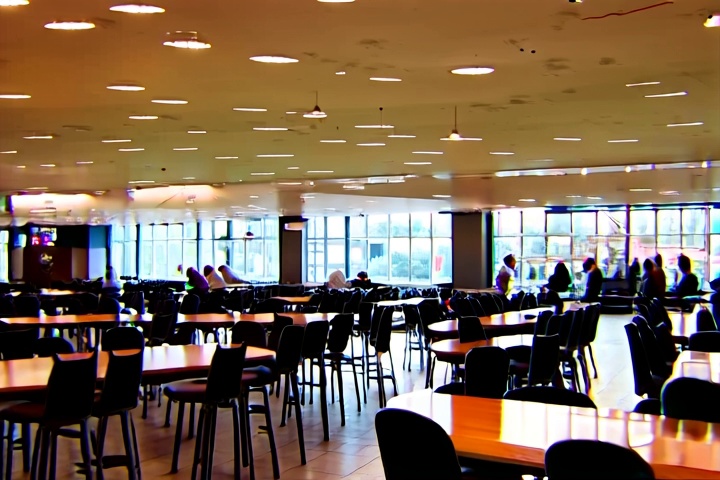} & 
        \includegraphics[width=75px, height=55px]{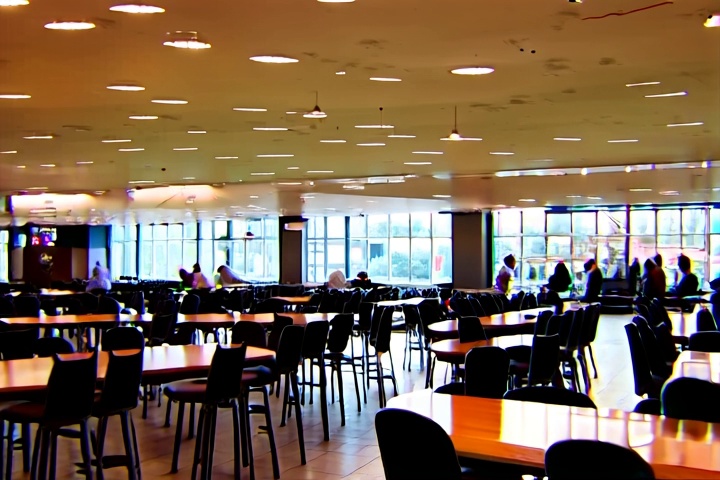} & 
        \includegraphics[width=75px, height=55px]{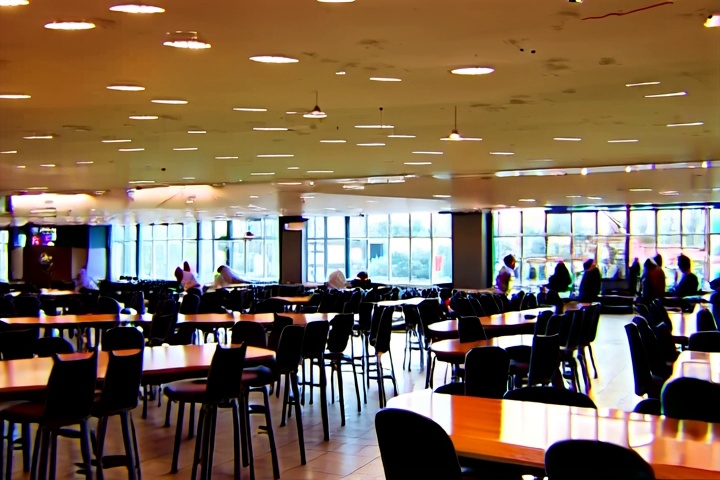} \\ 

        &
        \multicolumn{3}{c}{\small \textsf{``A cute happy Corgi playing in park''}} &
        \multicolumn{3}{c}{\small \textsf{``{Food} court''}} \\
    \end{tabular}
    }
    \vspace{-10pt}
    \caption{\textbf{Intra-frame qualitative visualization.} 
    \method enables the model to generate videos with improved image quality 
    and stronger semantic alignment. 
    \textit{Left}: Comparison of Image Quality. 
    It avoids generating objects with strange colors and reduces 
    visual artifacts (\eg, in the dog case), while producing harmonious objects (\eg, fire hydrants, vases). 
    \textit{Right}: Comparison of Semantic Alignment. 
    It generates correct scenes (\eg, restaurant), 
    accurate character relationships (\eg, couple), and proper visual elements (\eg, boat).
    }
    \vspace{-10pt}
    \label{fig:qualityin}
\end{figure*}

\cref{fig:qualityin} presents qualitative comparisons achieved using \method on baseline models. 
Videos generated after \method demonstrate enhanced visual details with fewer artifacts (as shown in the Quality column) 
and improved alignment with the input prompt (as shown in the Semantic column).
For instance, Turbo-\method produces a more accurate vase compared to the original model, 
where the vase’s mouth is incorrectly shaped. 
Similarly, the video from CogVideo-\method is better than that by CogVideo 
which contains unnatural color artifacts in the dog. 
In terms of semantic accuracy, 
VC2-\method generates a boat with visible human figures, offering a more accurate depiction compared to 
both VC2-VADER and the original VC2. 
Additionally, Turbo and CogVideo after training with our method, 
each generates more realistic character relationships and scene layouts correspondingly.
These improvements demonstrate that our alignment approach successfully enhances 
both semantic following and visual fidelity in generated videos.

\paragraph{Inter-frame qualitative analysis. }

\begin{figure*}[htbp]
    \centering
    \resizebox{1.0\textwidth}{!}{
    \begin{tabular}{p{2pt}c@{}c@{}cc@{}c@{}c}
        \multicolumn{4}{c}{\emph{Before Alignment}} & \multicolumn{3}{c}{\emph{After Alignment}} \\
        \multirow{1}{*}[33px]{\rotatebox{90}{VC2~\cite{chen2024videocrafter2}}} & 
        \includegraphics[width=75px, height=55px]{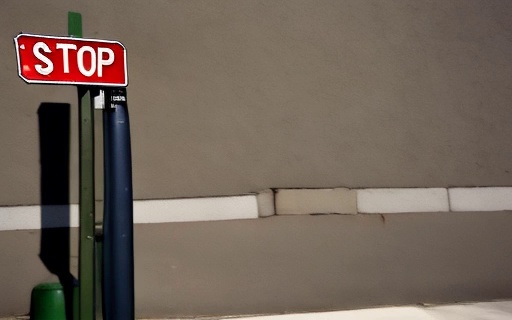} &
        \includegraphics[width=75px, height=55px]{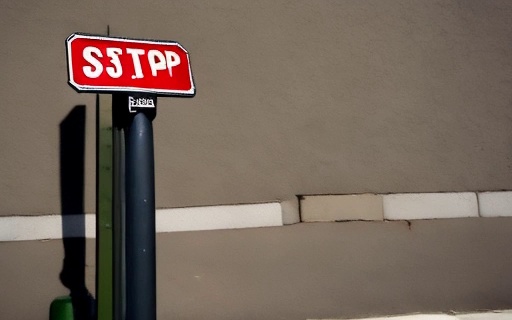} & 
        \includegraphics[width=75px, height=55px]{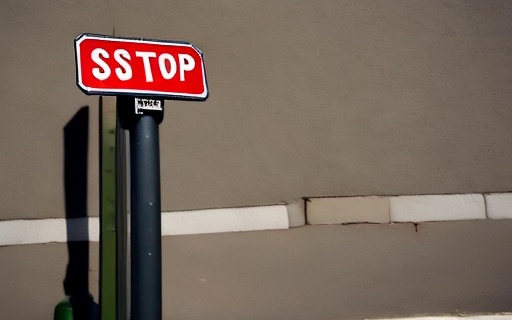} & 
        \includegraphics[width=75px, height=55px]{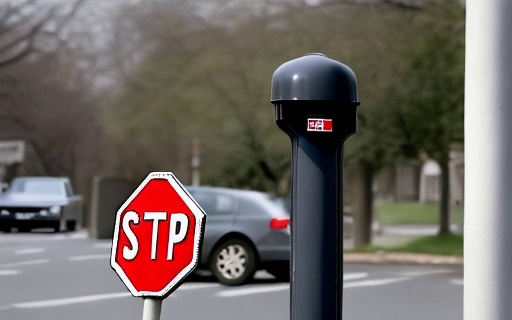} & 
        \includegraphics[width=75px, height=55px]{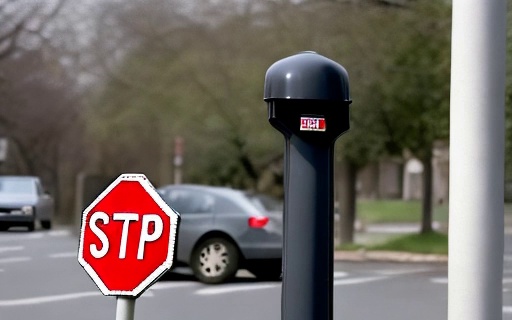} & 
        \includegraphics[width=75px, height=55px]{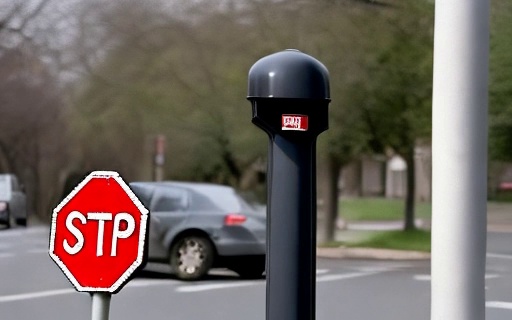} \\ 
        &
        \multicolumn{6}{c}{\small \textsf{``A stop sign and a parking meter''}} \\
        \multirow{1}{*}[38px]{\rotatebox{90}{Turbo~\cite{li2024t2vturbo}}} & 
        \includegraphics[width=75px, height=55px]{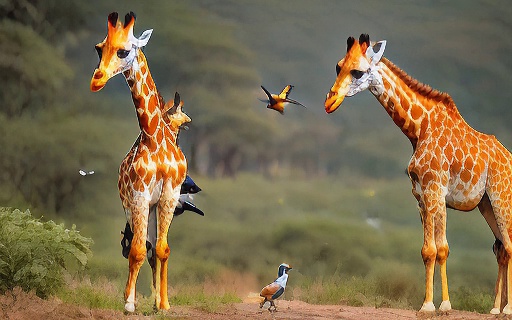} & 
        \includegraphics[width=75px, height=55px]{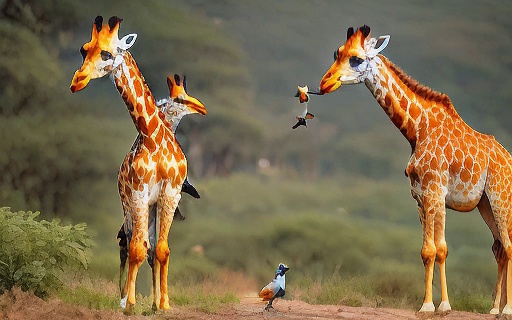} & 
        \includegraphics[width=75px, height=55px]{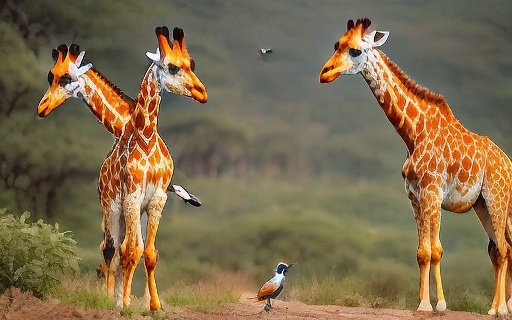} & 
        \includegraphics[width=75px, height=55px]{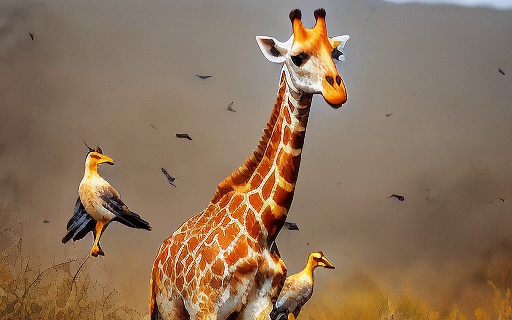} & 
        \includegraphics[width=75px, height=55px]{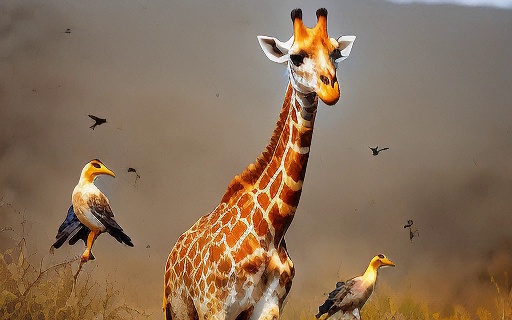} & 
        \includegraphics[width=75px, height=55px]{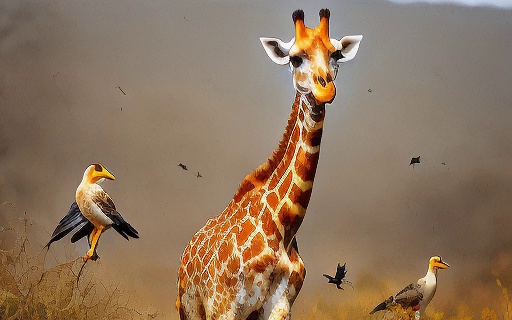} \\ 
        &
        \multicolumn{6}{c}{\small \textsf{``A giraffe and birds''}} \\
        \multirow{1}{*}[40px]{\rotatebox{90}{CogV.~\cite{hong2022cogvideo}}} & 
        \includegraphics[width=75px, height=55px]{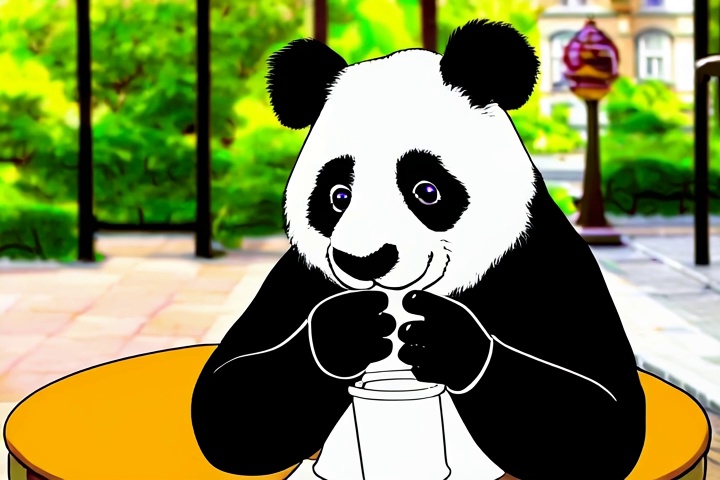} & 
        \includegraphics[width=75px, height=55px]{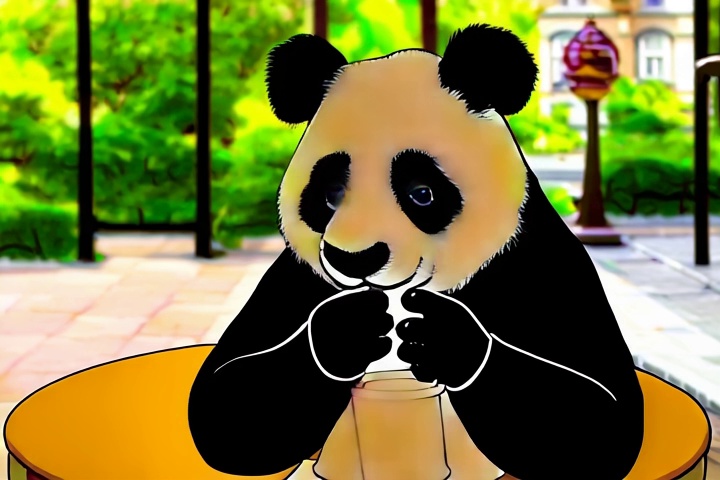}& 
        \includegraphics[width=75px, height=55px]{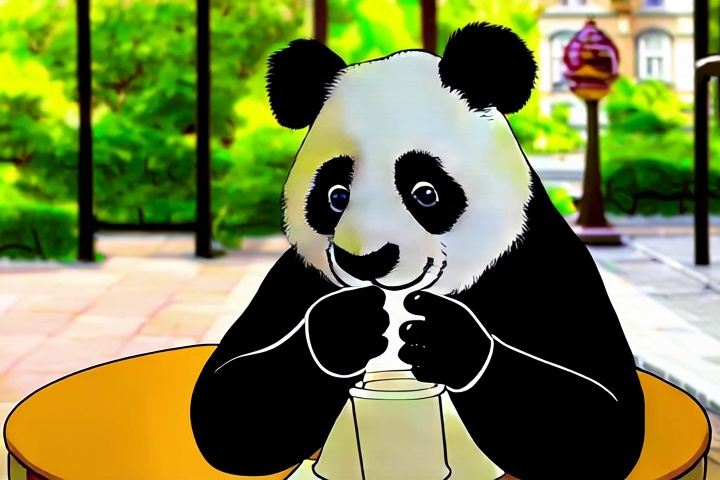}& 
        \includegraphics[width=75px, height=55px]{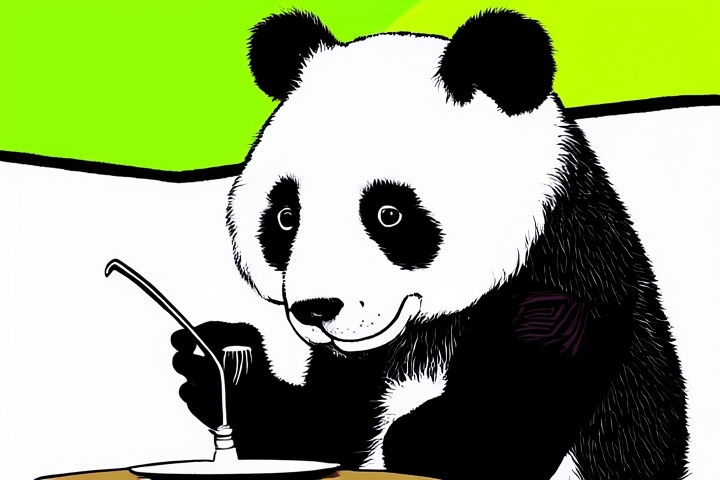}& 
        \includegraphics[width=75px, height=55px]{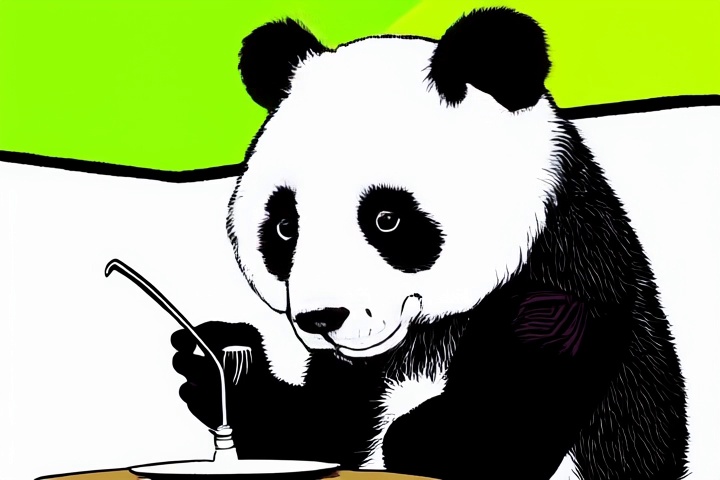}& 
        \includegraphics[width=75px, height=55px]{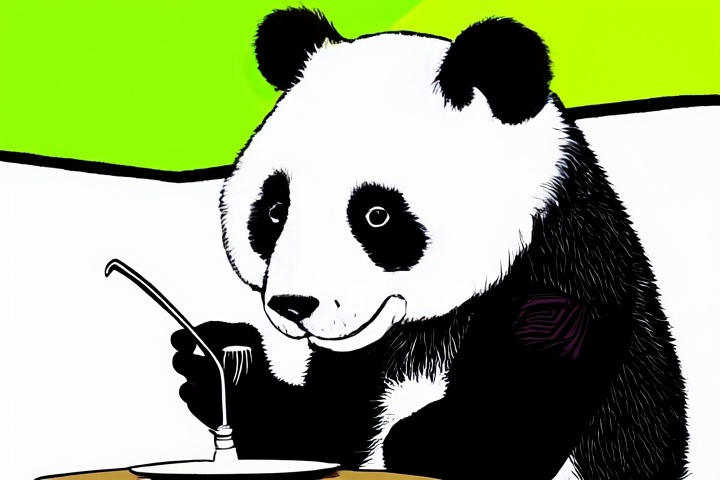} \\ 
        &
        \multicolumn{6}{c}{\small \textsf{``A panda''}} \\
    \end{tabular}
    }
    \vspace{-10pt}
    \caption{\textbf{Inter-frame qualitative visualization.}
    \method improves inter-frame consistency quality. 
    The aligned model generates stable text like in the signboard, 
    maintains consistent object appearances such as the giraffe, 
    and ensures uniform color tones, as seen with pandas.
    }
    \vspace{-5pt}
    \label{fig:qualityinter}
\end{figure*}

Our approach significantly improves the temporal consistency of aligned models,
 and \cref{fig:qualityinter} shows the comparison results. 
After alignment, VC2 is able to generate a stable stop sign, 
Turbo produces a scene where the number of giraffes remains consistent, 
and CogVideo generates a panda with stable coloring, avoiding sudden color changes. 
These examples demonstrate the effectiveness of our alignment method in enhancing temporal stability across frames, 
in terms of texts, object and color across different frames.

\subsection{Analysis}

\paragraph{Comparing \scorename with single-aspect reward. }
We compare the setting trained using \scorename with those trained using a single reward, 
such as  only the semantic score or the aesthetic score. On VBench, 
the results are 80.20\% and 79.65\%, which are significantly lower than 
the result achieved with \scorename, which is 81.93\%. This demonstrates the advantage 
of using a comprehensive reward like \scorename to evaluate samples. 
The comparison with VADER in \cref{tab:main}, 
also supports this conclusion.

\paragraph{Pairwise training strategies. }
\label{exp:pair}

We explore different strategies for constructing preference pairs 
from \( N \) generated videos for a prompt, shown in \cref{tab:pair}. 
The ``Better-Worse'' strategy outputs multiple pairs \( (v_i, v_j) \) 
as long as \( s_i > s_j \), representing the score of video \( v_i \) is higher than that of \( v_j \). 
The ``Best-vs-Worse'' strategy forms pairs by selecting the highest-scoring video 
and pairing it with others that have lower scores. 
In contrast, the ``Better-vs-Worst'' strategy pairs the lowest-scoring video with others. 
The ``Best-vs-Worst'' strategy, adopted by us, pairs only the highest 
and lowest-scoring videos and outputs 1 preference pair for a prompt. 
The experiments show that this strategy, which generates only the most distinctive pair, 
yields the best performance. 
This demonstrates that the key to the alignment performance 
lies in the average quality of the preference, rather than the absolute quantity of data.

\begin{table*}[ht]
    \small
    \centering
    \setlength{\tabcolsep}{3.8pt}
    \begin{subtable}[t]{0.5\textwidth}
        \centering
        \begin{tabular}{l c c c c c}
            \toprule
            \multirow{2}{*}{\textbf{Method}} & \multicolumn{3}{c}{\textbf{VBench (\%)}} & \multirow{2}{*}{\textbf{HPS (V)}} & \multirow{2}{*}{\textbf{PickScore}} \\
            \cmidrule(lr){2-4}
            & \textbf{Total} & \textbf{Q} & \textbf{S} & & \\
            \midrule
            better \vs worse & 81.32 & 83.46 & 72.74 & 0.258 & 20.62\\
            best \vs worse & 80.80 & 82.74 & 73.03 & 0.258 & 20.62\\
            better \vs worst & 80.73 & 82.40 & 74.08 & 0.259 & 20.67 \\
            \rowcolor[gray]{0.85}
            best \vs worst & \textbf{81.93} & 83.07 & 77.38 & 0.261 & 20.65 \\
            \bottomrule
        \end{tabular}
        \caption{\small Performance on different pairing strategies.}
        \label{tab:pair}
    \end{subtable}%
    \setlength{\tabcolsep}{3pt}
    \begin{subtable}[t]{0.45\textwidth}
        \centering
        \begin{tabular}{l c c c c c}
            \toprule
            \multirow{2}{*}{\textbf{Method}} & \multicolumn{3}{c}{\textbf{VBench (\%)}} & \multirow{2}{*}{\textbf{HPS (V)}} & \multirow{2}{*}{\textbf{PickScore}} \\
            \cmidrule(lr){2-4}
            & \textbf{Total} & \textbf{Q} & \textbf{S} & & \\
            \midrule
            -75 & 80.08 & 81.53 & 74.29 & 0.259 & 20.64\\
            -50 & 81.29 & 82.94 & 74.68 & 0.259 & 20.57\\
            -25 & 81.42 & 83.16 & 74.49 & 0.258 & 20.65\\
            \rowcolor[gray]{0.85}
            Full & \textbf{81.93} & 83.07 & 77.38 & 0.261 & 20.65\\
            \bottomrule
        \end{tabular}
        \caption{\small Filtering out the least distinct pairs at different ratios. }
        \label{tab:filter}
    \end{subtable}

    \vspace{-10pt}
    \vskip\baselineskip
    \setlength{\tabcolsep}{3pt}
    \begin{subtable}[t]{0.5\textwidth}
        \centering
        \begin{tabular}{l c c c c c}
            \toprule
            \multirow{2}{*}{\textbf{$\alpha$}} & \multicolumn{3}{c}{\textbf{VBench (\%)}} & \multirow{2}{*}{\textbf{HPS (V)}} & \multirow{2}{*}{\textbf{PickScore}} \\
            \cmidrule(lr){2-4}
             & \textbf{Total} & \textbf{Q} & \textbf{S} & & \\
            \midrule
            0.0 \textit{Vanilla DPO} & 80.89 & 82.78 & 73.32 & 0.260 & 20.64\\
            0.5 & 81.51 & 82.99 & 75.60 & 0.260 & 20.68\\
            \rowcolor[gray]{0.85}
            1.0 \textit{Ours} & \textbf{81.93} & 83.07 & 77.38 & 0.261 & 20.65 \\
            2.0 & 81.93 & 82.52 & 79.59 & 0.260 & 20.70\\
            \bottomrule
        \end{tabular}
        \caption{\small Different values of $\alpha$.}
        \label{tab:reweight}
    \end{subtable}%
    \setlength{\tabcolsep}{4.8pt}
    \begin{subtable}[t]{0.45\textwidth}
        \centering
        \begin{tabular}{l c c c c c}
            \toprule
            \multirow{2}{*}{\textbf{$N$}} & \multicolumn{3}{c}{\textbf{VBench (\%)}} & \multirow{2}{*}{\textbf{HPS (V)}} & \multirow{2}{*}{\textbf{PickScore}} \\
            \cmidrule(lr){2-4}
             & \textbf{Total} & \textbf{Q} & \textbf{S} & & \\
            \midrule
            2 & 80.89 & 82.78 & 73.32 & 0.260 & 20.60 \\
            3 & 81.51 & 82.99 & 75.60 & 0.260 & 20.62 \\
            \rowcolor[gray]{0.85}
            4 & \textbf{81.93} & 83.07 & 77.38 & 0.261 & 20.65 \\
            \bottomrule
            \\
        \end{tabular}
        \caption{\small Different values of $N$.}
        \label{tab:N}
    \end{subtable}
    \vspace{-5pt}
    \caption{\textbf{Ablation studies.} We study different strategies and configurations including (a) the pair strategy, 
    (b) the filter strategy, (c) $\alpha$ values,  the tuning hyper-parameter for re-weighting, and (d) $N$ values, the number of video samples for each text prompt. \textbf{Q} is short for visual quality and \textbf{S} is short for semantic alignment.}
    \label{tab:combined}
    \vspace{-15pt}
\end{table*}

\paragraph{Data filtering.}

According to \cref{fig:distribution}, there are some preference pairs are not distinctive, 
that the positive video is only slightly better than the negative one in terms of the \scorename. 
We explored the impact of removing these less-distinctive pairs to see if it could improve alignment performance. 
However, the results in \cref{tab:filter} show that excluding these training samples led to worse performance. 
This may be because the removal of these pairs, also including the prompts, reduced the diversity of the training data, 
leaving the model with many prompts it had not seen before and weakening its alignment performance.
Though the pairs distinctiveness of these prompts are small, 
they still play a crucial role in the alignment.

\paragraph{Preference re-weighting scale. }
Determining the scale of re-weighting for distinctive preference pairs is important. 
Here, we explore the impact of different values of \( \alpha \) on alignment learning performance. 
When \( \alpha = 0 \), all pairs are assigned equal weight, disabling the re-weighting mechanism as a vanilla DPO. 
A higher \( \alpha \) value increases the weight assigned to rare preference pairs.
As shown in \cref{tab:reweight}, 
a value of \( \alpha = 1 \) performs significantly better than \( \alpha = 0.5 \), 
providing a reference for the model's weight scaling. 
At \( \alpha = 2 \), we observe a significant increase in semantic performance 
but a decrease in quality performance on VBench, 
resulting in the same total score as achieved with \( \alpha = 1 \). 
All non-zero values of \( \alpha \) improve performance, 
demonstrating the robustness of the re-weighting approach.

\paragraph{Comparing with supervised fine-tuning. }
We compare \method with supervised fine-tuning (SFT), 
a baseline approach of post-training for pre-trained models. 
In SFT, only winning samples \( v^W \) are used to fine-tune the model, 
while all other settings remain the same.
\cref{tab:main} shows for models like VC2 and CogVideo, 
\method shows a clear advantage over SFT. 
This suggests the importance of learning from negative samples, 
which reduces the likelihood of generating lower-quality outputs.

\paragraph{Effect of varying $N$ on performance. }
We investigate the impact of generating different numbers of videos $N$ on model performance. 
As shown in \cref{fig:distribution}, a larger $N$ tends to produce more distinctive samples. 
\cref{tab:N} presents the experimental results for varying values of $N$, 
indicating that performance improves as $N$ increases. 
However, a larger $N$ also increases the cost of dataset generation. 
Determining the optimal balance between performance gain 
and data generation cost as $N$ grows remains an open question.

\vspace{-2px}
\section{Conclusion}
\vspace{-6px}
In this paper, we propose \method, a novel pipeline to align video diffusion models.
\method introduces a comprehensive scoring method, \scorename, along with 
a novel data reweighting strategy that automatically constructs and prioritizes preference data, 
enabling more effective alignment training.
Experiments show that \method enhances both visual quality 
and semantic alignment for state-of-the-art text-to-video models. 

{
    \small
    \bibliographystyle{ieeenat_fullname}
    \bibliography{main}
}

\clearpage
\setcounter{page}{1}
\maketitlesupplementary
\appendix

This supplementary material presents \scorename details, additional analysis and experimental results. 
Section~\ref{supp:score} enumerates the details of \scorename, including the model each dimenson ultilizes and the corresponding weights. 
Section~\ref{supp:analysis} compares the performance of single-dimensional, multi-dimensional settings and  aggregation methods, also examines the impact of training data scale on the results.
Section~\ref{supp:qual} includes additional intra-frame and inter-frame qualitative results.

\section{\scorename Implementation.}\label{supp:score}
Inspired by the models used in \cite{huang2023vbench}, we build OmniScore by referencing these models and their corresponding weights to evaluate the quality of video samples. \cite{huang2023vbench} aims to evaluate the quality of video generative models, whereas our \scorename targets assessing the quality of video samples specifically for preference learning. 
Here we demonstrate the detailed composition of \scorename:

\paragraph{Motion Smoothness.} We utilize the motion priors in the video frame interpolation model \cite{licvpr23amt} to evaluate the smoothness of generated motions
\paragraph{Temporal Flickering.} We take static frames by RAFT \cite{teed2020raft} and compute the mean absolute difference across frames. 
\paragraph{Subject Consistency}- For a subject(e.g., a person, a car, or a cat) in the video, we assess whether its appearance remains consistent throughout the whole video. To this end, we calculate the DINO \cite{caron2021emerging} feature similarity across frames.
\paragraph{Imaging Quality}. Imaging quality refers to the distortion (e.g., over-exposure, noise, blur)presented in the generated frames, and we evaluate it using the MUSIQ \cite{Ke2021MUSIQ} image quality predictor trained on the SPAQ \cite{Fang2020spaq} dataset.
\paragraph{Aesthetic Quality}. We evaluate the
 artistic and beauty value perceived by humans towards each video frame using the LAION aesthetic predictor \cite{LAIONaes}.
\paragraph{Dynamic Degree} We use RAFT \cite{teed2020raft} to estimate the degree of dynamics in synthesized videos.
\paragraph{Text-Video semantic alignment.} We use overall video-text consistency computed by ViCLIP \cite{wang2023internvid}.

The following dimensions are scaled to the range $[0,1]$ based on the following values:  
\begin{itemize}
    \item \textbf{Subject Consistency}: $\text{Min} = 0.1462, \text{Max} = 1.0$
    \item \textbf{Temporal Flickering}: $\text{Min} = 0.6293, \text{Max} = 1.0$
    \item \textbf{Motion Smoothness}: $\text{Min} = 0.706, \text{Max} = 0.9975$
    \item \textbf{Overall Consistency}: $\text{Min} = 0.0, \text{Max} = 0.364$
\end{itemize}

The weights assigned to Motion Smoothness, Temporal Flickering, Subject Consistency, Imaging Quality, Aesthetic Quality and Dynamic Degree are all 4, and the weight for Text-Video Semantic Alignment is set to 1. 

\section{Additional Analysis}\label{supp:analysis}
\paragraph{Single- vs. multi-dimensional score comparison.}
In Table~\ref{tab:train-objectives}, we explore the results of training on a single-dimensional 
reward score compared to training on our \scorename. The experimental 
results show that \scorename achieves the best performance, 
highlighting the importance of a comprehensive score for our framework.

\paragraph{Multi-dimensional score aggregation.}
We explore two methods for multi-dimensional score aggregation: (1) selecting 10,000 pairs based on our \scorename and (2) Combine preference pairs from individual dimensions into a larger dataset so that the VC2 model is trained on 40,000 pairs, with 10,000 pairs selected from each of the four dimensions: semantics, aesthetics, motion smoothness, and dynamic degree. The results indicate that the second approach significantly lowers performance to 78.26\% on VBench-Total, showing that using our \scorename can achieve better performance.

\paragraph{Effect of training scale on performance.}
We compared the performance shown in Table~\ref{tab:dsize} when using only half and 25\% of the prompt data for training, 
observing a significant drop across all metrics. 
This result demonstrates that increasing the amount of prompt data in training 
yields substantially better performance. We attribute this to improved generalization, 
as the model aligns with a broader range of prompts. 
These experiments suggest that our method still has room for improvement, 
particularly with regard to the amount of data.

\begin{table}[ht]
    \centering
    \resizebox{0.5\textwidth}{!}{
    \resizebox{\linewidth}{!}{
        \begin{tabular}{l c c c c c}
            \toprule
            \multirow{2}{*}{\textbf{Data}} & \multicolumn{3}{c}{\textbf{VBench(\%)}} & \multirow{2}{*}{\textbf{HPS (V)}} & \multirow{2}{*}{\textbf{PickScore}} \\
            \cmidrule(lr){2-4}
             & \textbf{Total} & \textbf{Quality} & \textbf{Semantic} & & \\
            \midrule
            25\% & 80.21 & 81.70 & 74.26 & 0.259 & 20.66\\
            50\% & 80.83 & 82.37 & 74.68 & 0.260 & 20.59\\
            \rowcolor[gray]{0.85}
            Full(ours) & {81.93} & 83.07 & 77.38 & 0.261 & 20.65\\
            \bottomrule
        \end{tabular}
    }
    }
    \caption{Scores for Different Dataset Sizes}
    \label{tab:dsize}
    \vspace{-6px}
\end{table}

\section{Additional Qualitative Results}\label{supp:qual}
\begin{figure*}[htbp]
    \centering
    \begin{tabular}{p{2pt}c@{}c@{}cc@{}c@{}c}
        \multicolumn{4}{c}{\emph{Before Alignment}} & \multicolumn{3}{c}{\emph{After Alignment}} \\
        \multirow{1}{*}[33px]{\rotatebox{90}{VC2~\cite{chen2024videocrafter2}}} & 
        \includegraphics[width=75px, height=55px]{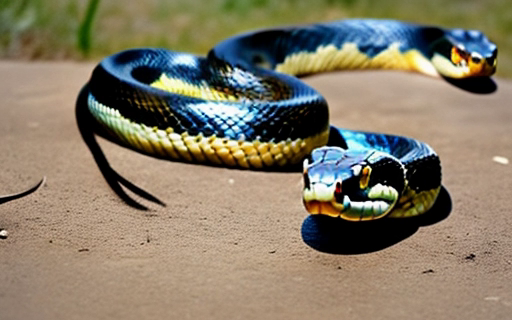} &
        \includegraphics[width=75px, height=55px]{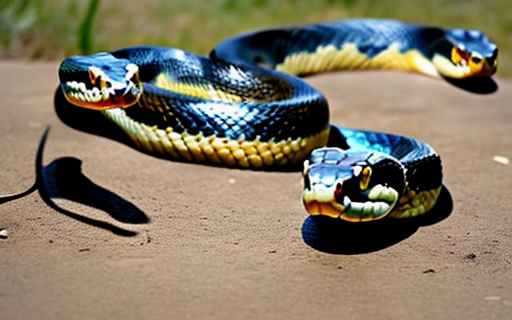} & 
        \includegraphics[width=75px, height=55px]{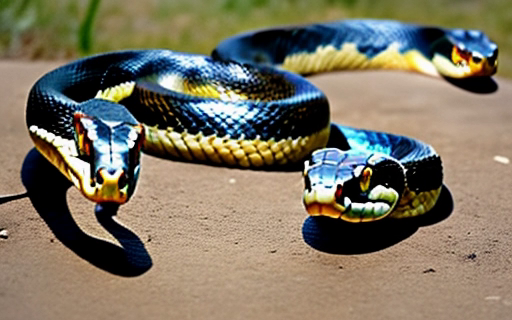} & 
        \includegraphics[width=75px, height=55px]{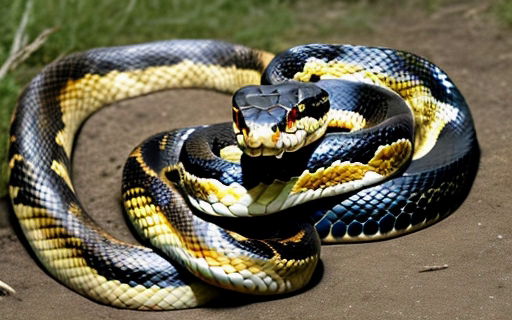} & 
        \includegraphics[width=75px, height=55px]{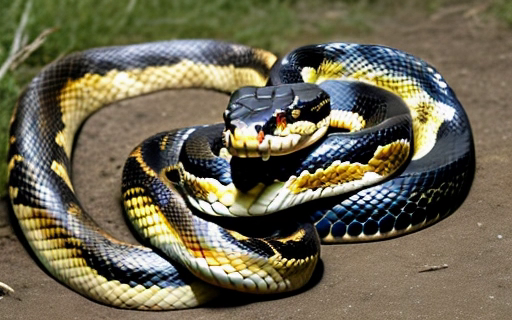} & 
        \includegraphics[width=75px, height=55px]{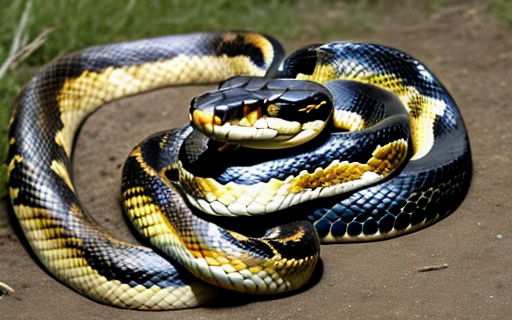} \\ 
        &
        \multicolumn{6}{c}{\small \textsf{``Pythons, for example, can engage in cannibalism''}} \\
        \multirow{1}{*}[33px]{\rotatebox{90}{VC2~\cite{chen2024videocrafter2}}} & 
        \includegraphics[width=75px, height=55px]{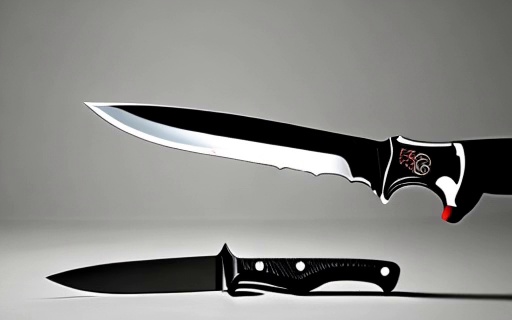} &
        \includegraphics[width=75px, height=55px]{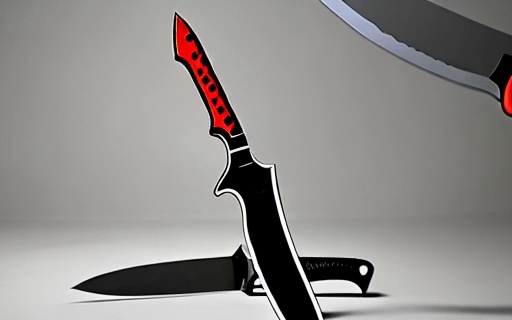} & 
        \includegraphics[width=75px, height=55px]{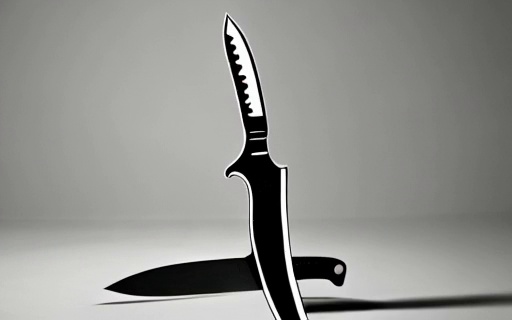} & 
        \includegraphics[width=75px, height=55px]{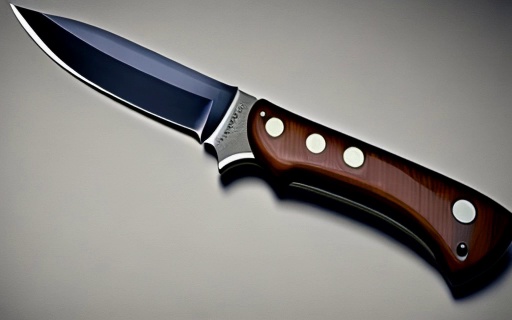} & 
        \includegraphics[width=75px, height=55px]{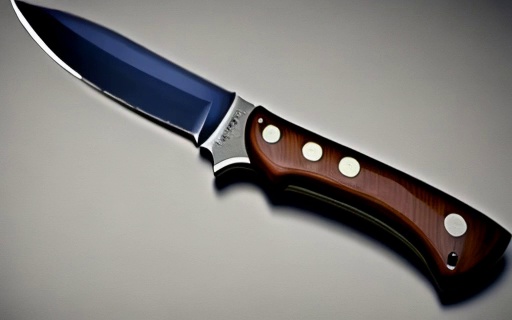} & 
        \includegraphics[width=75px, height=55px]{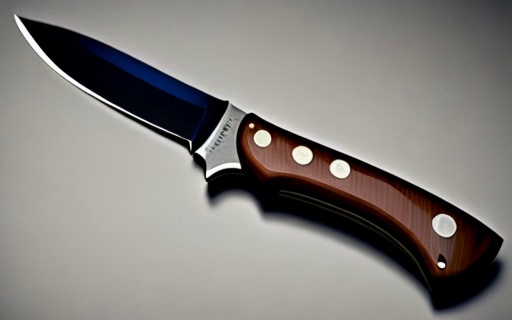} \\ 
        &
        \multicolumn{6}{c}{\small \textsf{``A knife''}} \\
        \midrule
        \multirow{1}{*}[38px]{\rotatebox{90}{Turbo~\cite{li2024t2vturbo}}} & 
        \includegraphics[width=75px, height=55px]{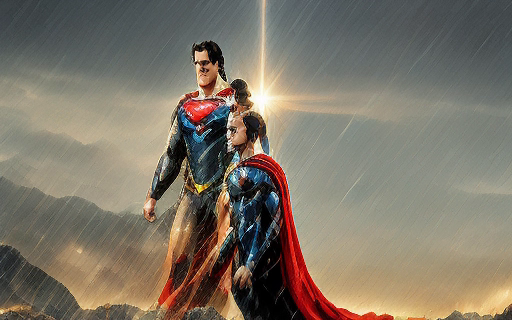} &
        \includegraphics[width=75px, height=55px]{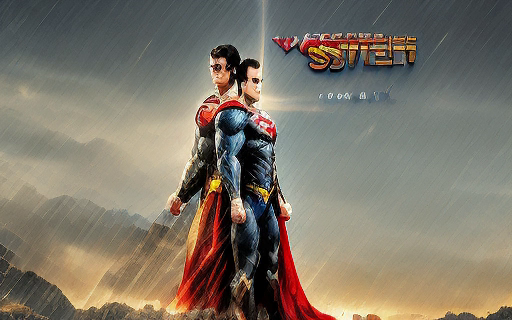} & 
        \includegraphics[width=75px, height=55px]{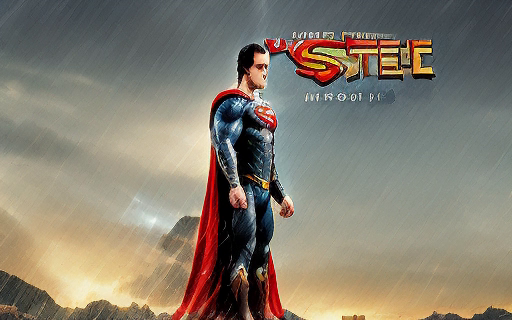} & 
        \includegraphics[width=75px, height=55px]{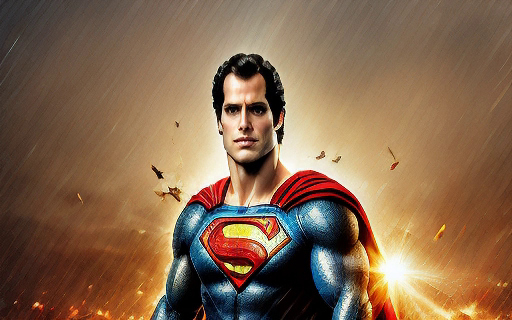} & 
        \includegraphics[width=75px, height=55px]{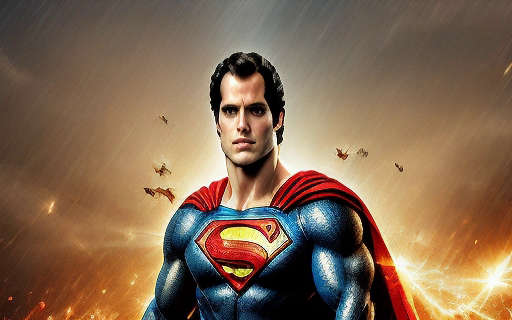} & 
        \includegraphics[width=75px, height=55px]{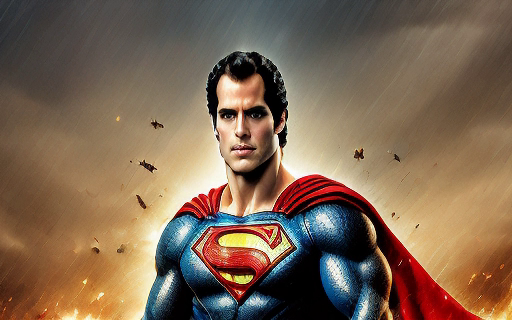} \\ 
        &
        \multicolumn{6}{c}{\small \textsf{``Space man playing instruments''}} \\
        \multirow{1}{*}[38px]{\rotatebox{90}{Turbo~\cite{li2024t2vturbo}}} & 
        \includegraphics[width=75px, height=55px]{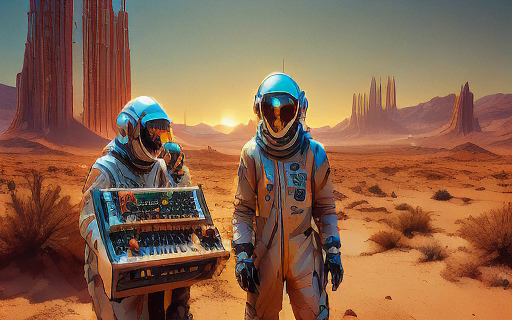} &
        \includegraphics[width=75px, height=55px]{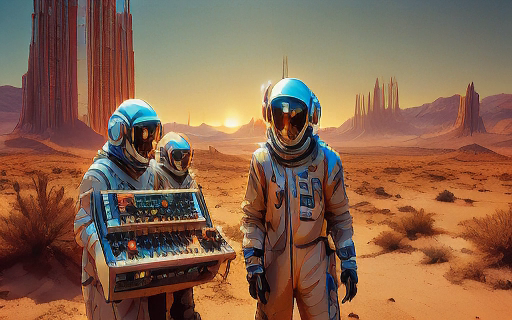} & 
        \includegraphics[width=75px, height=55px]{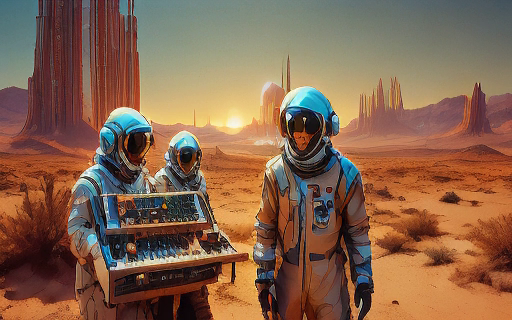} & 
        \includegraphics[width=75px, height=55px]{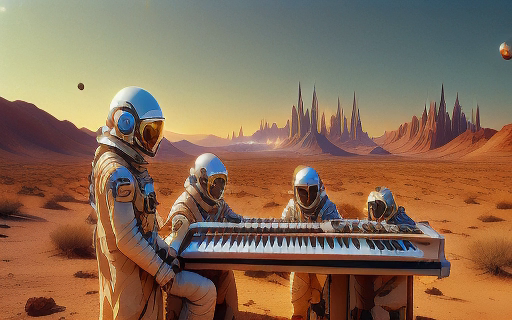} & 
        \includegraphics[width=75px, height=55px]{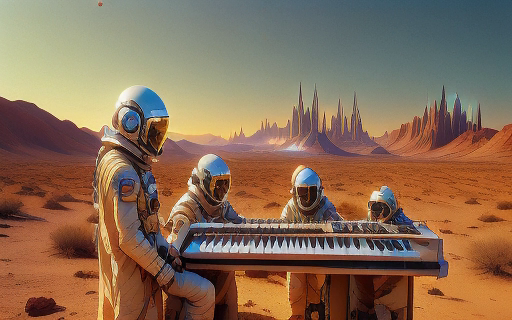} & 
        \includegraphics[width=75px, height=55px]{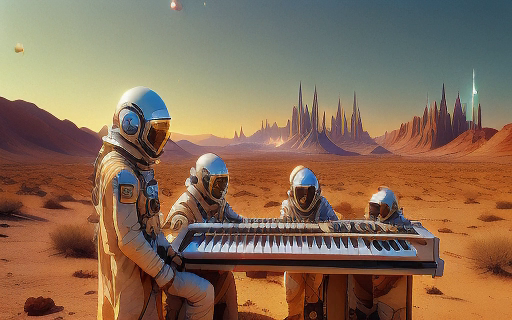} \\ 
        &
        \multicolumn{6}{c}{\small \textsf{``cinematic Man of steel movie poster''}} \\
        \midrule
        \multirow{1}{*}[38px]{\rotatebox{90}{CogV.~\cite{li2024t2vturbo}}} & 
        \includegraphics[width=75px, height=55px]{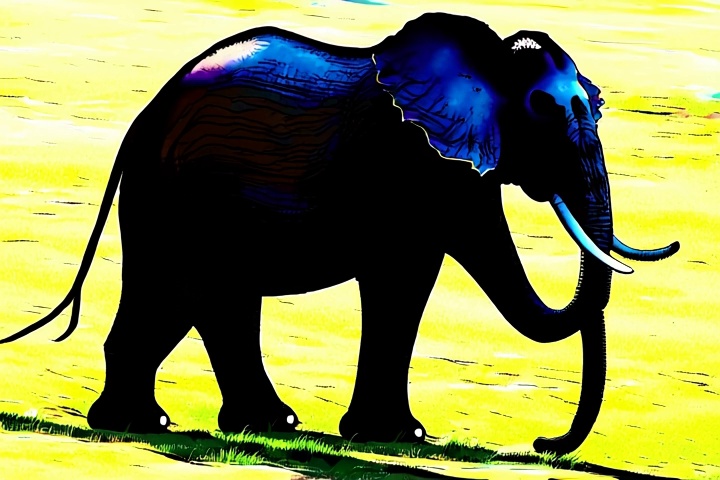} &
        \includegraphics[width=75px, height=55px]{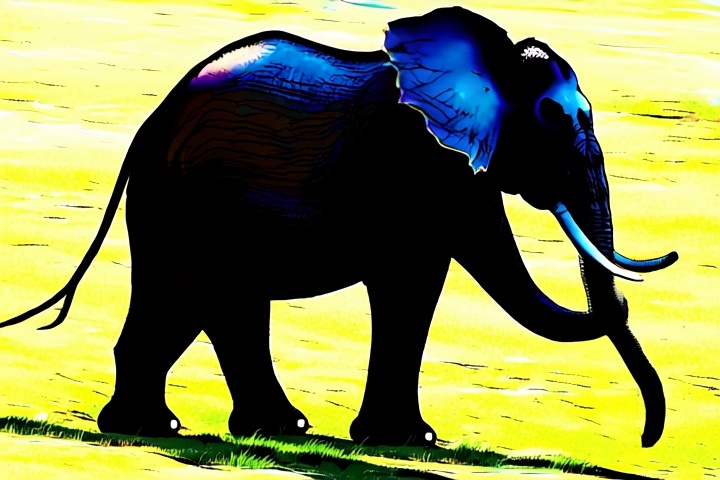} & 
        \includegraphics[width=75px, height=55px]{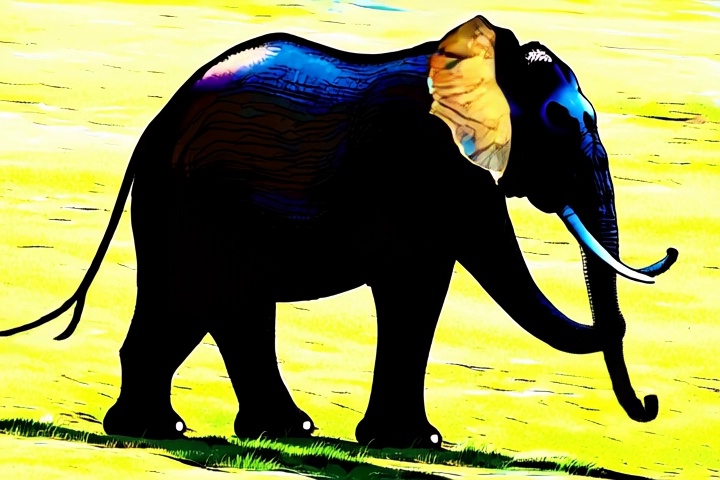} & 
        \includegraphics[width=75px, height=55px]{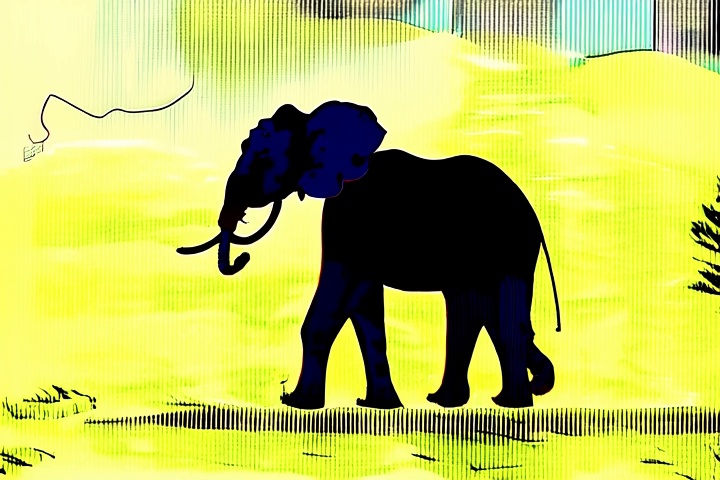} & 
        \includegraphics[width=75px, height=55px]{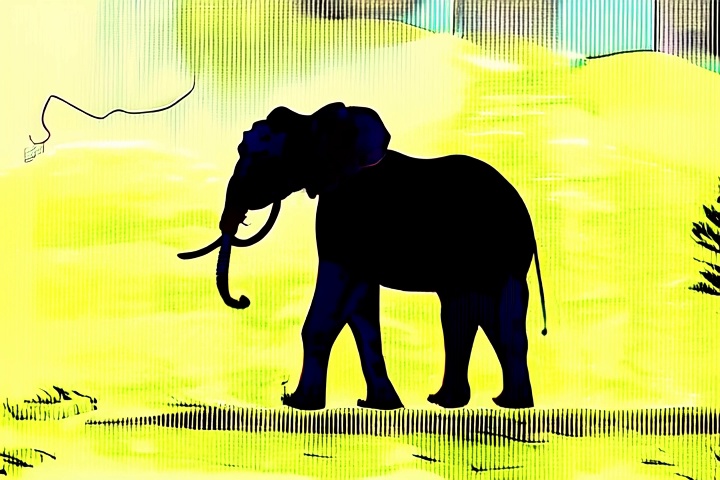} & 
        \includegraphics[width=75px, height=55px]{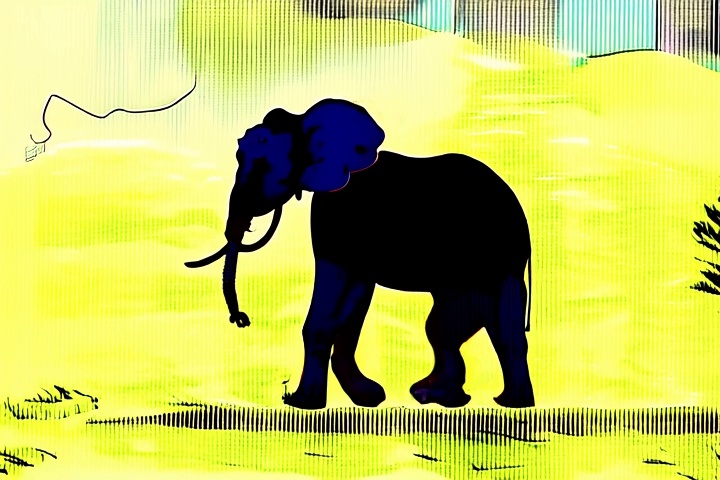} \\
        &
        \multicolumn{6}{c}{\small \textsf{``an elephant taking a peaceful walk''}} \\
        \multirow{1}{*}[38px]{\rotatebox{90}{CogV.~\cite{li2024t2vturbo}}} & 
        \includegraphics[width=75px, height=55px]{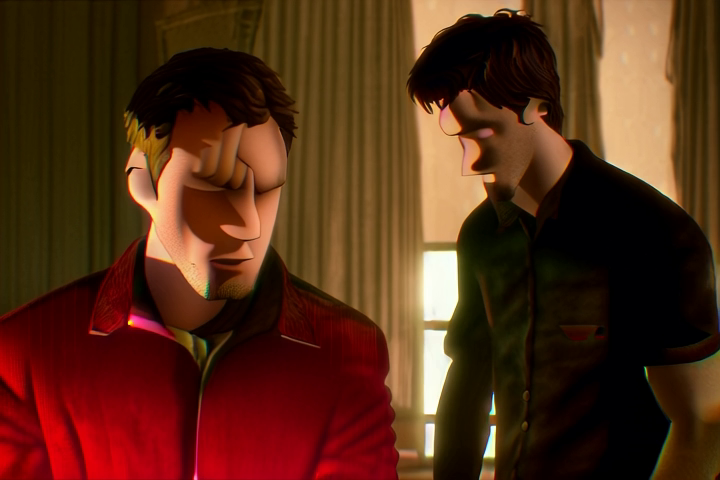} &
        \includegraphics[width=75px, height=55px]{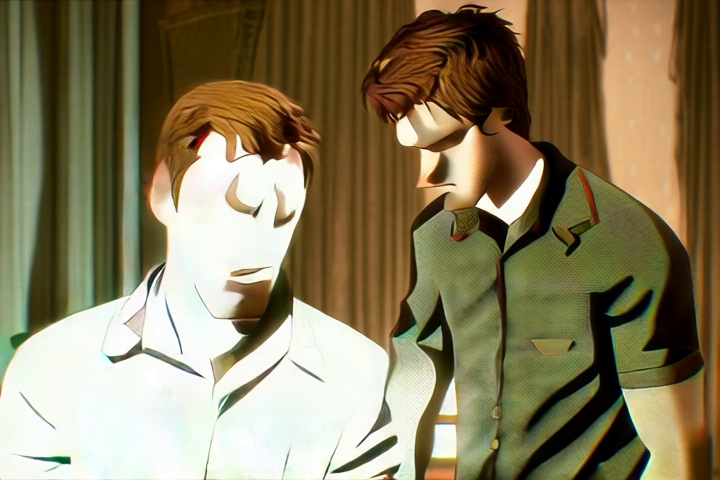} &
        \includegraphics[width=75px, height=55px]{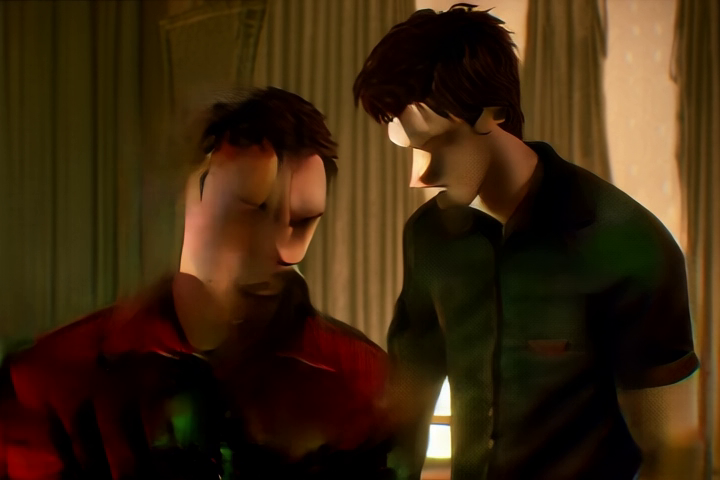} &
        \includegraphics[width=75px, height=55px]{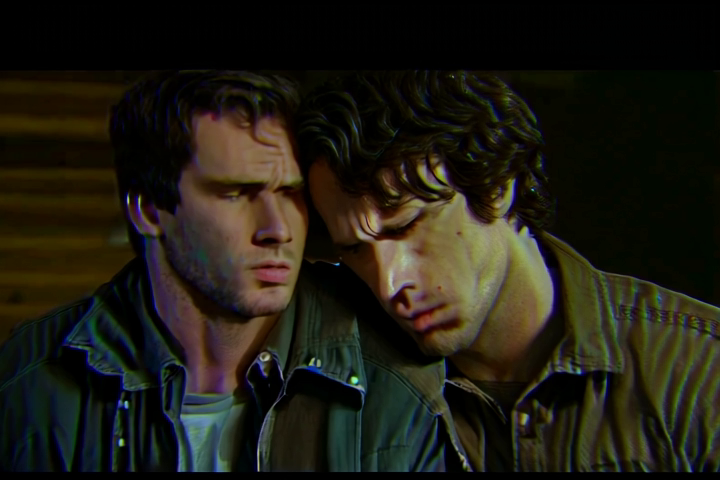} &
        \includegraphics[width=75px, height=55px]{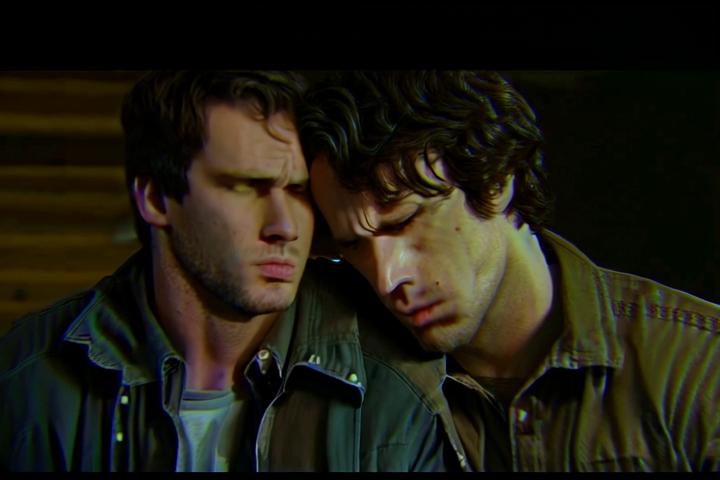} &
        \includegraphics[width=75px, height=55px]{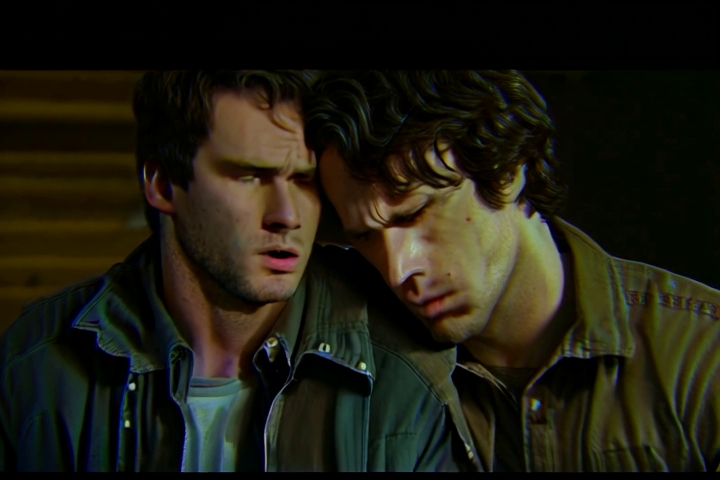} \\ 
        &
        \multicolumn{6}{c}{\small \textsf{``Two mans are talking''}} \\
        
    \end{tabular}
    \vspace{-10pt}
    \caption{Additional inter-frame qualitative visualization.
    }
    \vspace{-5pt}
    \label{suppfig:qualityinter}
\end{figure*}

\begin{figure*}[htbp]
    \centering
    \resizebox{1.0\textwidth}{!}{
    \begin{tabular}{p{2pt}c@{}c@{}cc@{}c@{}c}
        \multicolumn{4}{c}{\emph{Visual Quality}} & \multicolumn{3}{c}{\emph{Semantic Alignment}} \\
        \multirow{1}{*}[35px]{\rotatebox{90}{VC2~\cite{chen2024videocrafter2}}} & 
        \includegraphics[width=75px, height=55px]{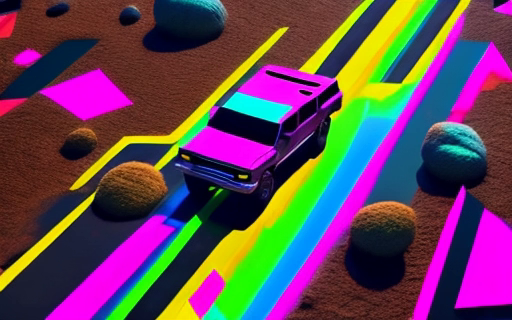} & 
        \includegraphics[width=75px, height=55px]{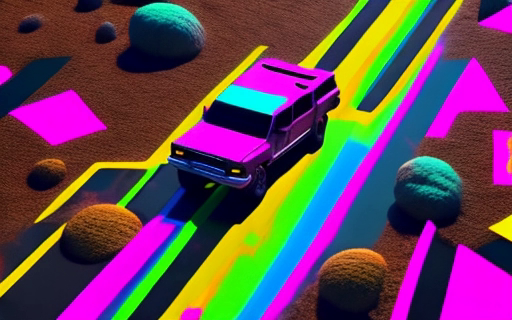} & 
        \includegraphics[width=75px, height=55px]{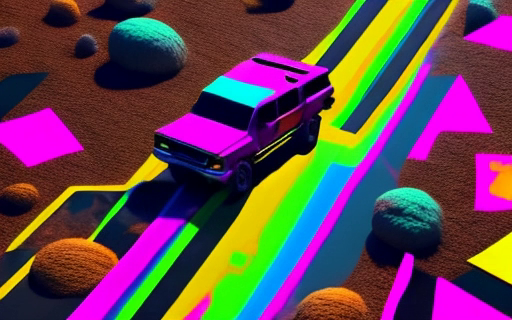} & 
        \includegraphics[width=75px, height=55px]{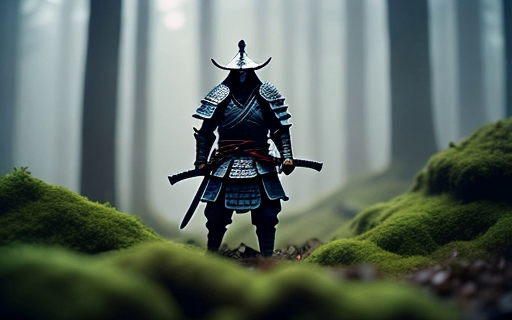} & 
        \includegraphics[width=75px, height=55px]{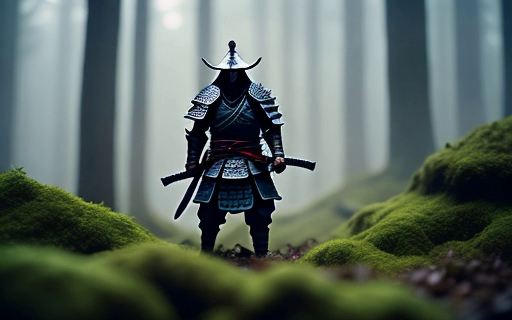} & 
        \includegraphics[width=75px, height=55px]{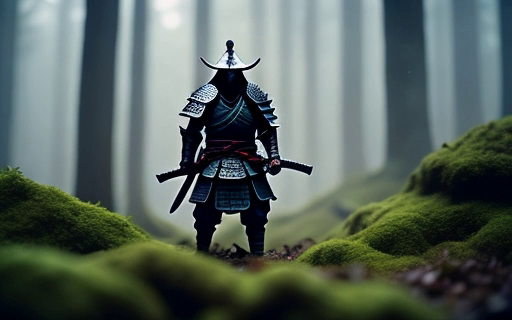} \\ 
        \multirow{1}{*}[45px]{\rotatebox{90}{VADER~\cite{prabhudesai2024VADER}}} & 
        \includegraphics[width=75px, height=55px]{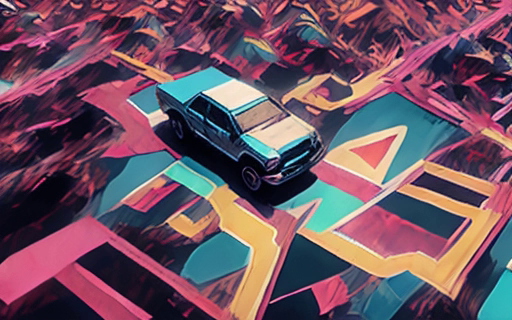} & 
        \includegraphics[width=75px, height=55px]{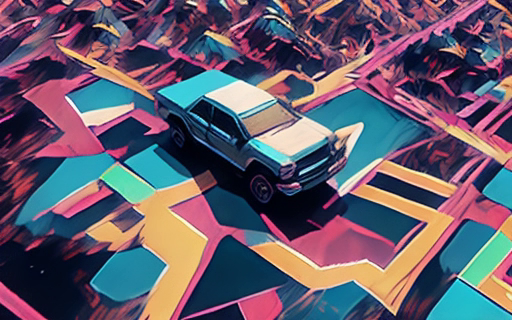} & 
        \includegraphics[width=75px, height=55px]{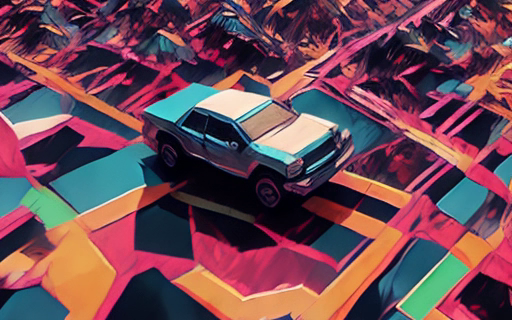} & 
        \includegraphics[width=75px, height=55px]{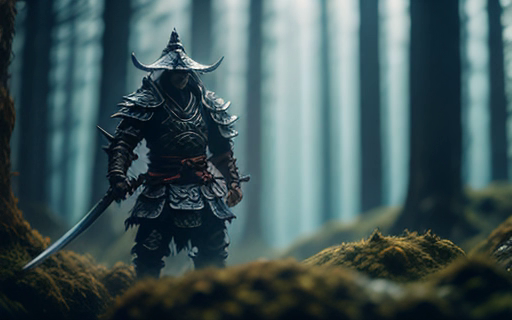} & 
        \includegraphics[width=75px, height=55px]{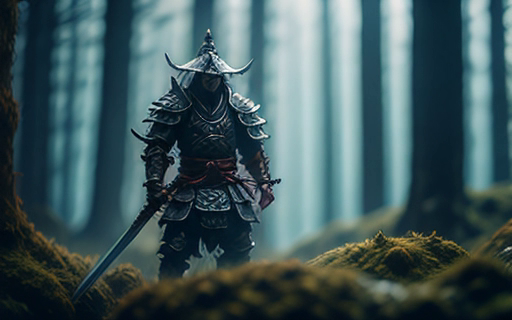} & 
        \includegraphics[width=75px, height=55px]{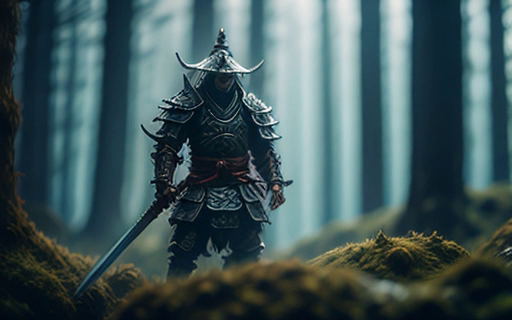}\\ 
        \multirow{1}{*}[40px]{\rotatebox{90}{\method}} & 
        \includegraphics[width=75px, height=55px]{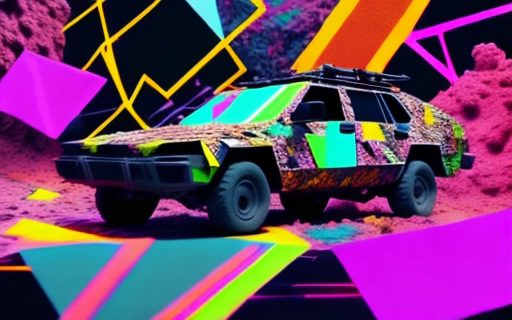} & 
        \includegraphics[width=75px, height=55px]{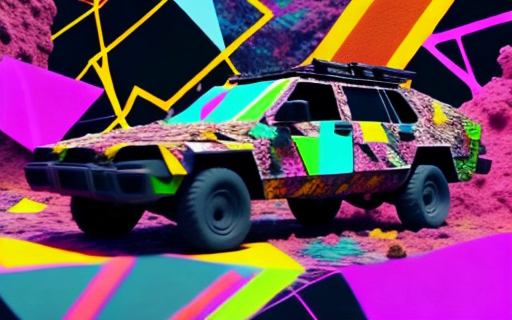} & 
        \includegraphics[width=75px, height=55px]{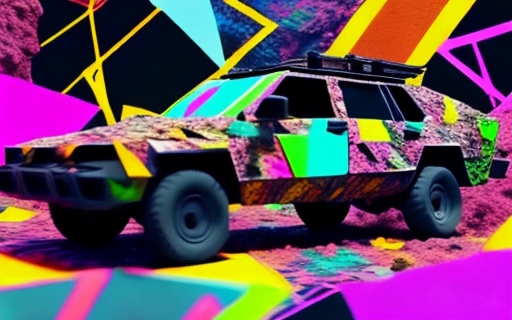} & 
        \includegraphics[width=75px, height=55px]{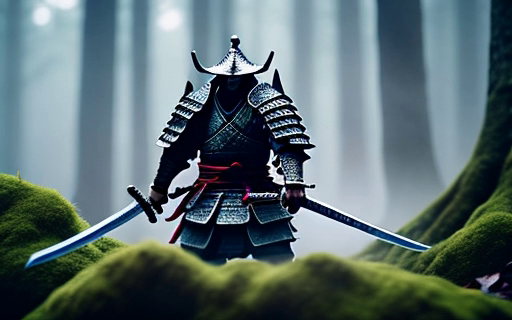} & 
        \includegraphics[width=75px, height=55px]{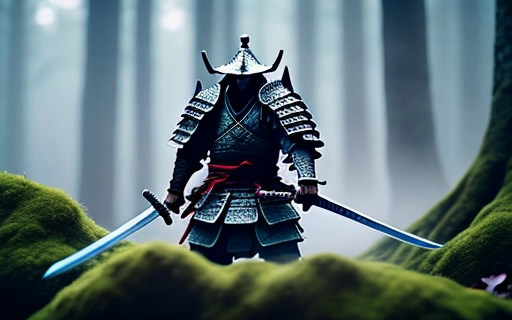} & 
        \includegraphics[width=75px, height=55px]{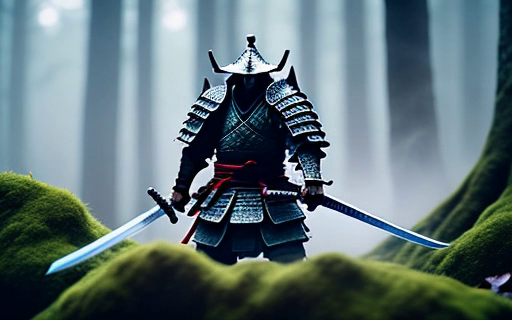} \\ 
        &
        \multicolumn{3}{c}{\small \textsf{``A camo-wrapped vehicle through an abstract landscape''}} &
        \multicolumn{3}{c}{\small \textsf{``samurai, armor, sword, mist, fog, moss, forest''}} \\

        \multirow{1}{*}[40px]{\rotatebox{90}{Turbo~\cite{li2024t2vturbo}}} & 
        \includegraphics[width=75px, height=55px]{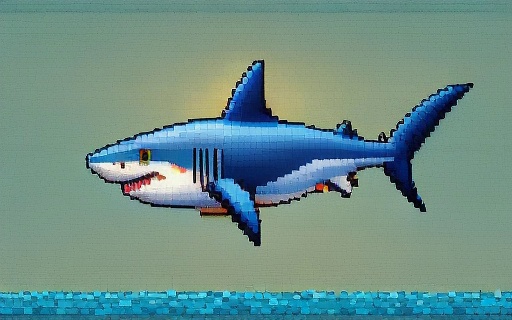} & 
        \includegraphics[width=75px, height=55px]{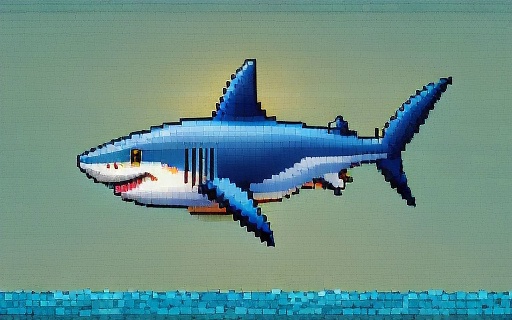} & 
        \includegraphics[width=75px, height=55px]{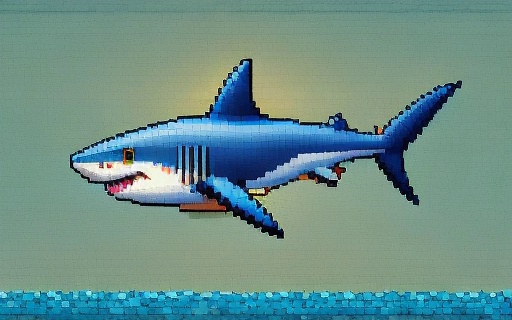} & 
        \includegraphics[width=75px, height=55px]{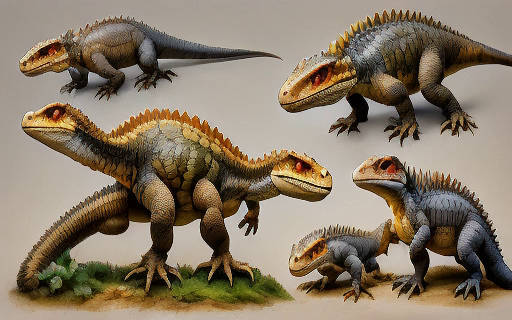} & 
        \includegraphics[width=75px, height=55px]{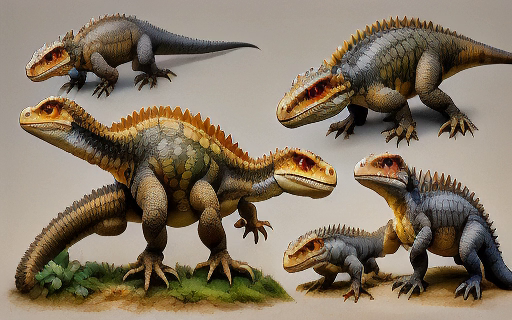} & 
        \includegraphics[width=75px, height=55px]{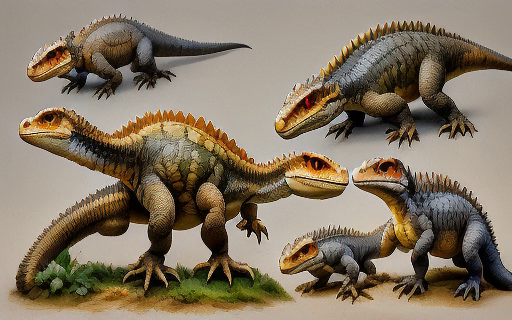}  \\ 
        \multirow{1}{*}[40px]{\rotatebox{90}{\method}} & 
        \includegraphics[width=75px, height=55px]{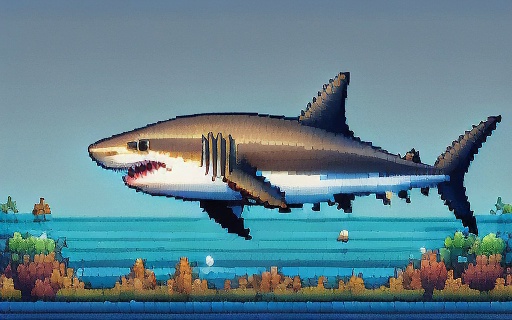} & 
        \includegraphics[width=75px, height=55px]{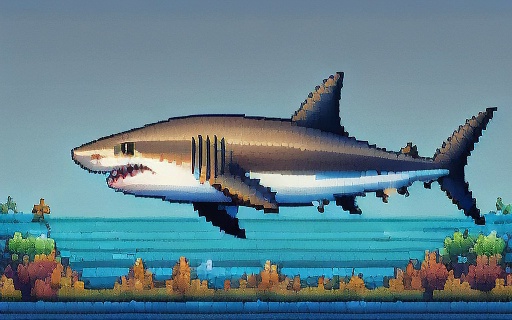} & 
        \includegraphics[width=75px, height=55px]{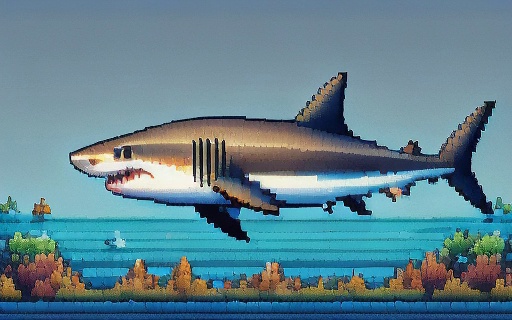} & 
        \includegraphics[width=75px, height=55px]{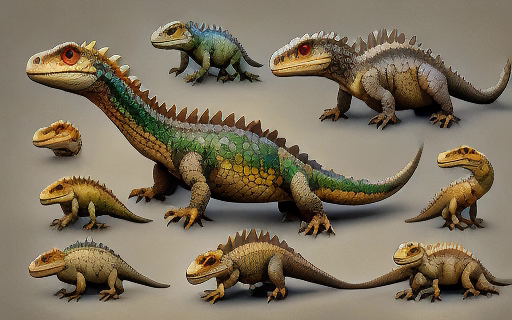} & 
        \includegraphics[width=75px, height=55px]{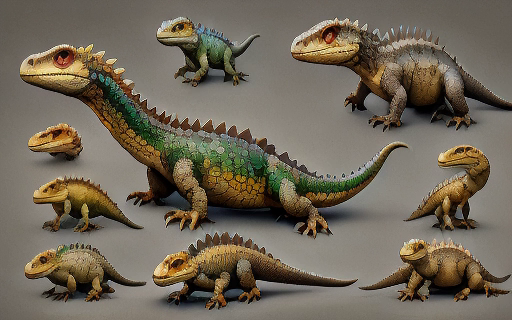} & 
        \includegraphics[width=75px, height=55px]{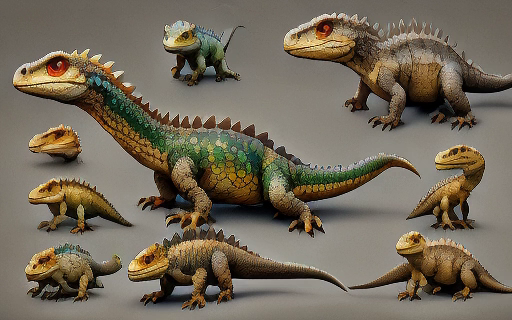} \\ 
        
        &
        \multicolumn{3}{c}{\small \textsf{``A shark is swimming in the ocean, pix art''}} &
        \multicolumn{3}{c}{\small \textsf{``Mesozoic Era different types of reptiles''}} \\
        \multirow{1}{*}[38px]{\rotatebox{90}{CogV.~\cite{hong2022cogvideo}}} & 
        \includegraphics[width=75px, height=55px]{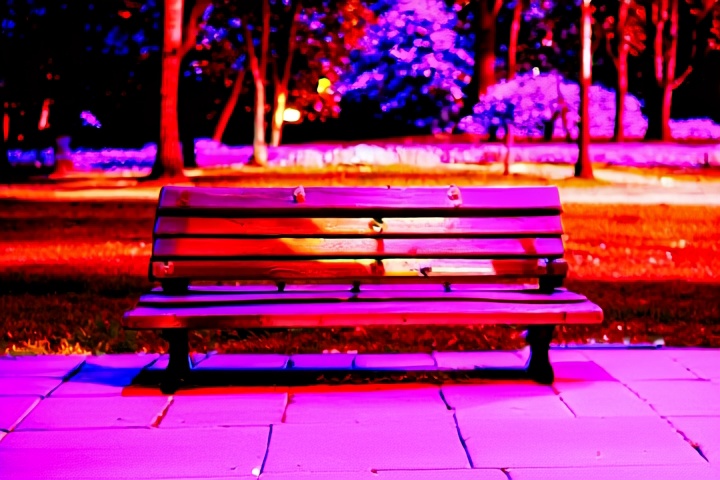} & 
        \includegraphics[width=75px, height=55px]{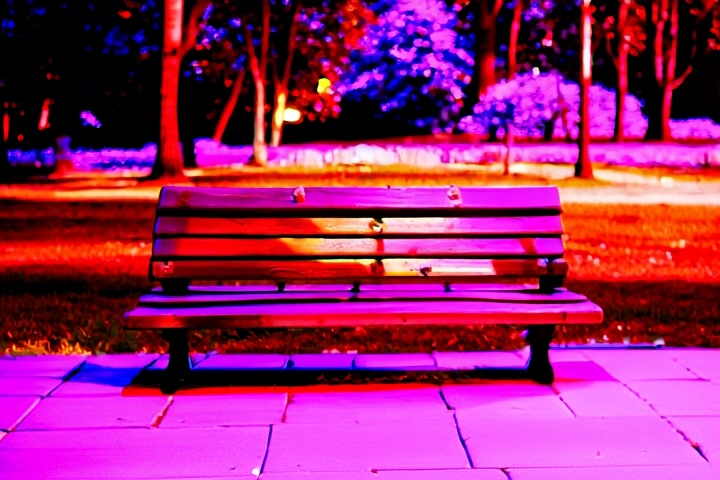} & 
        \includegraphics[width=75px, height=55px]{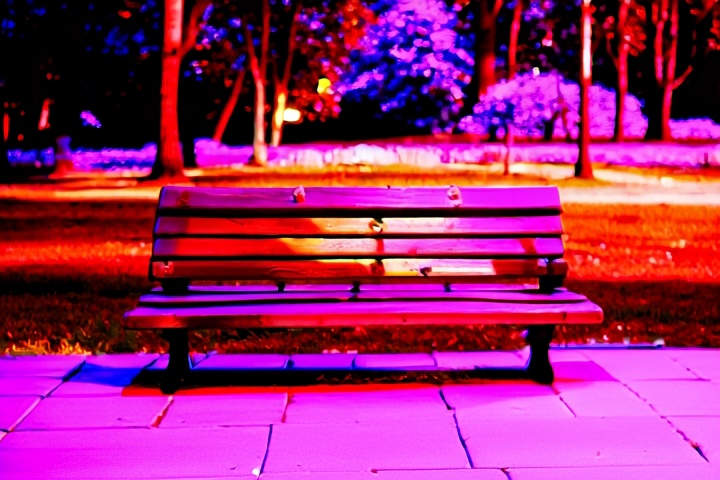} &
        \includegraphics[width=75px, height=55px]{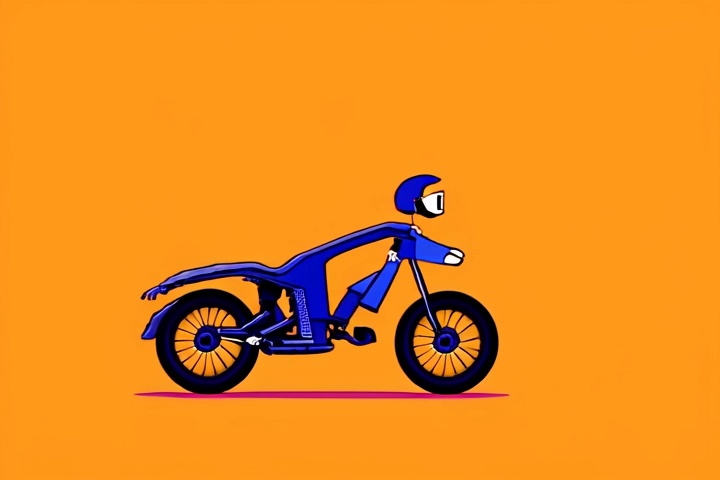} & 
        \includegraphics[width=75px, height=55px]{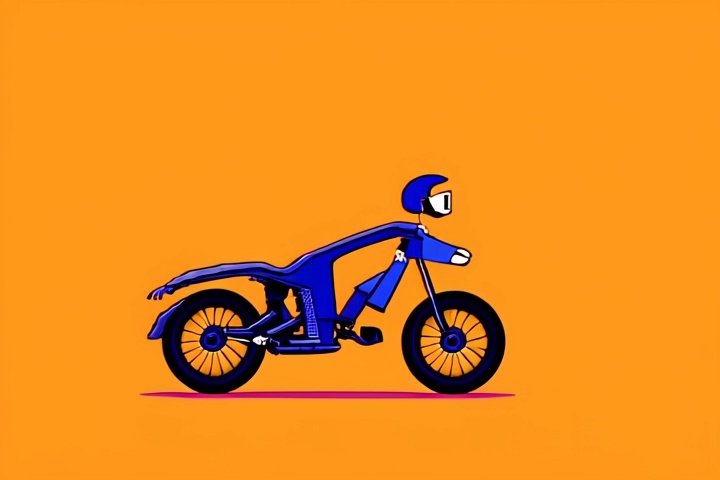} & 
        \includegraphics[width=75px, height=55px]{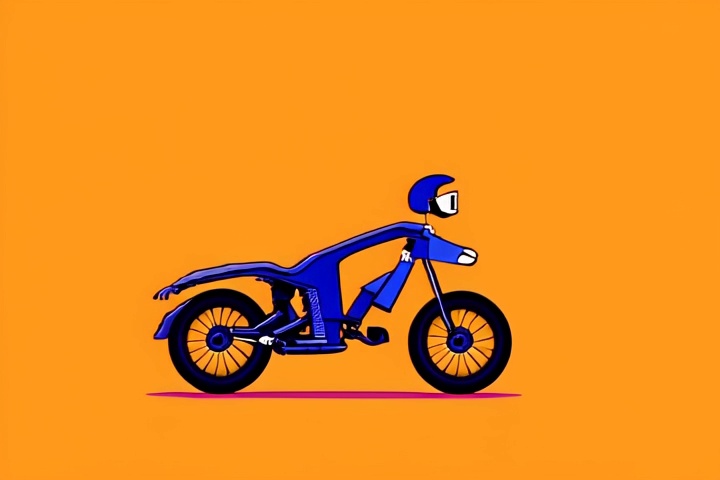} \\ 

        \multirow{1}{*}[40px]{\rotatebox{90}{\method}} & 
        \includegraphics[width=75px, height=55px]{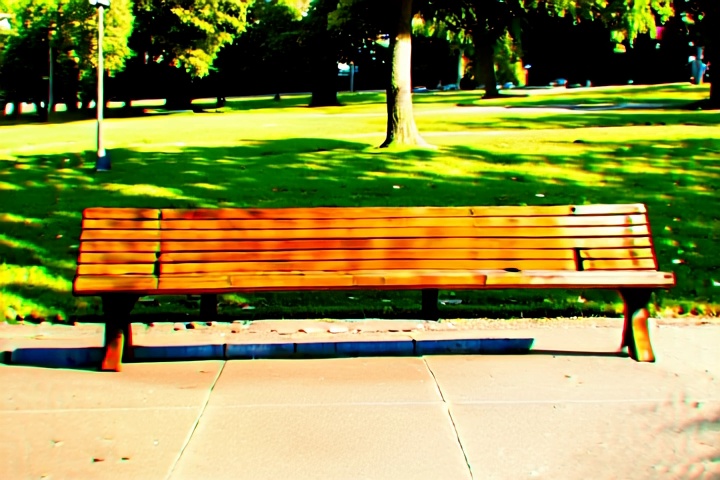} & 
        \includegraphics[width=75px, height=55px]{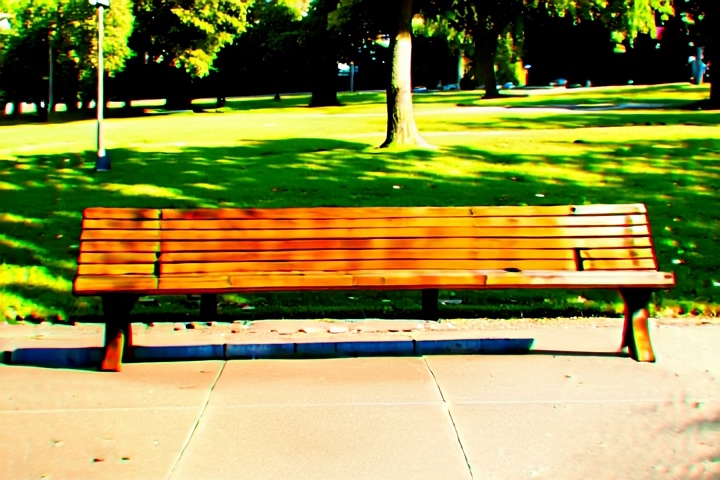} & 
        \includegraphics[width=75px, height=55px]{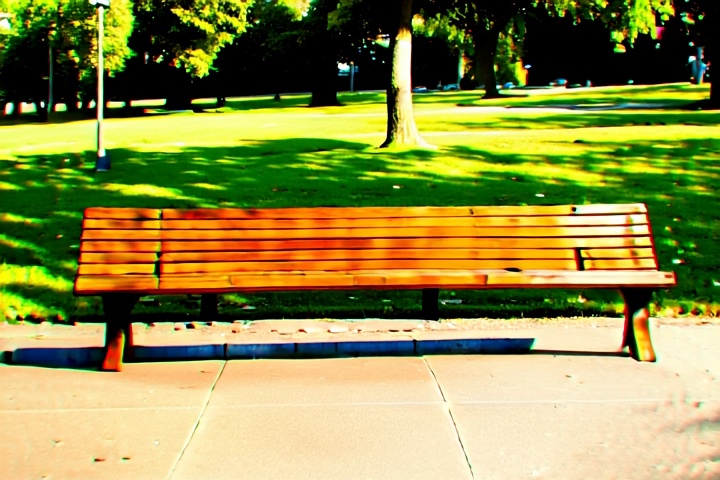} &
        \includegraphics[width=75px, height=55px]{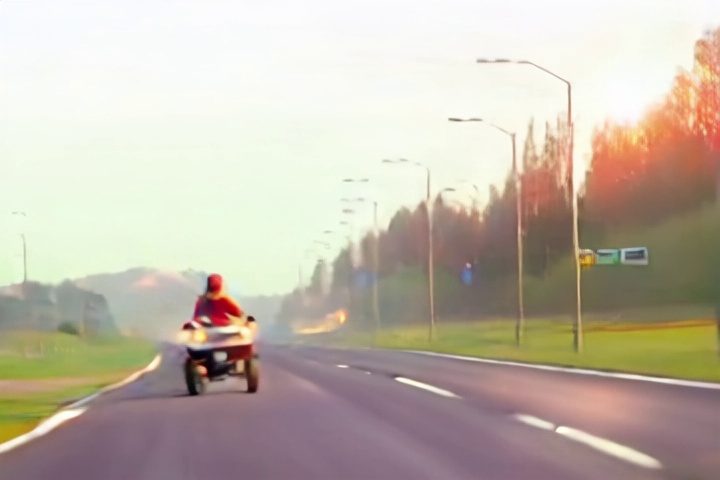} & 
        \includegraphics[width=75px, height=55px]{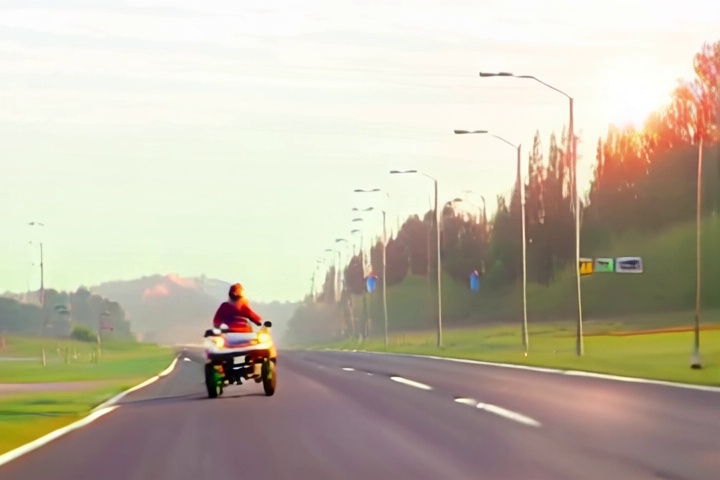} & 
        \includegraphics[width=75px, height=55px]{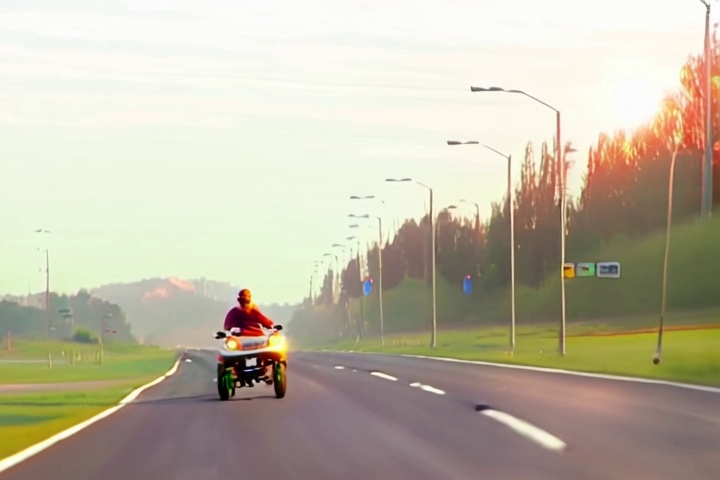} \\
        &
        \multicolumn{3}{c}{\small \textsf{``A tranquil tableau of a wooden bench in the park''}} &
        \multicolumn{3}{c}{\small \textsf{``A person is motorcycling''}} \\
    \end{tabular}
    }
    \vspace{-10pt}
    \caption{Additional intra-frame qualitative visualization.}
    \vspace{-10pt}
    \label{suppfig:qualityin}
\end{figure*}

We present the results of inter-frame and intra-frame alignment before and after learning in Figure~\ref{suppfig:qualityinter} and Figure~\ref{suppfig:qualityin}, respectively, following the format of the main paper. The results demonstrate that our alignment method is effective across a wide range of prompts, improving temporal consistency, visual quality, and semantics.

\begin{table*}[ht]
    \centering
        \begin{tabular}{l c c c c c c}
            \toprule
            \multirow{2}{*}{\textbf{Score}} & \multicolumn{3}{c}{\textbf{VBench (\%)}} & \multirow{2}{*}{\textbf{Subject Consis.}} & \multirow{2}{*}{\textbf{Aesthetic Quality}} & \multirow{2}{*}{\textbf{Overall Consis.}} \\
\cmidrule(lr){2-4}
 & \textbf{Total} & \textbf{Quality} & \textbf{Semantic} & & & \\
            \midrule
            Overall Consis. & 80.20 & 81.57 & 74.74 & 95.61 & 62.94 & 78.76 \\
            Aesthetic Quality   & 79.65 & 81.67 & 71.57 & 97.13 & 63.27 & 76.98 \\
            Subject Consis. & 77.05 & 79.00 & 69.28 & 94.25 & 58.23 & 73.35 \\
            \rowcolor[gray]{0.85} \scorename(ours) & 81.93 & 83.07 & 77.38 & 95.69 & 63.18 & 78.43 \\
            \bottomrule
        \end{tabular}
    \caption{Scores for different training objectives include single-dimensional scores such as overall consistency, aesthetic quality, and subject consistency, as well as our multi-dimensional score, OmniScore. "Consis." is the abbreviation for "consistency."}
    \label{tab:train-objectives}
\end{table*}

\end{document}